%% file: paper.tex
\documentclass[11pt]{mosi}
\input{source-preamble}
\title{DraMaze: A Multi-Agent Pipeline with Future-Aware Memory for Diverse Microdrama Script Generation}

\author{
  Siyang He$^{1,2,*}$\hspace{0.6em}
  Qiqi Wang$^{1,2,*}$\hspace{0.6em}
  Fan Liu$^{1}$\hspace{0.6em}
  Yuncheng Huang$^{3}$\\[0.5ex]
  Pengfei Xu$^{3}$\hspace{0.6em}
  Xi Chen$^{3,\dagger}$\hspace{0.6em}
  Ziwei He$^{2,\dagger}$\hspace{0.6em}
  Xipeng Qiu$^{1,2}$
}
\affiliation[1]{Fudan University}
\affiliation[2]{Shanghai Innovation Institute}
\affiliation[3]{Solar System of OVB, Tencent, China}
\contribution[*]{Equal contribution}
\contribution[\dagger]{Corresponding authors.}

\abstract{\input{abstract}}
\checkdata[Repository]{\url{https://github.com/rimory6/DraMaze.git}}

\correspondence{\email{ hsy2678501@gmail.com, qqwang7799@gmail.com}}

\begin{document}
\titleformat*{\paragraph}{\normalfont\bfseries}
\maketitle

\input{sections/01.tex}
\input{sections/02.tex}

\input{sections/03.tex}

\input{sections/04.tex}

\input{sections/05.tex}

\input{sections/06.tex}

\input{sections/07.tex}

\input{sections/08.tex}

\input{references.tex}

\input{appendices/01.tex}

\input{appendices/02.tex}

\input{appendices/03.tex}

\input{appendices/04.tex}

\input{appendices/05.tex}

\input{appendices/06.tex}
\input{appendices/07.tex}

\end{document}

%% file: source-preamble.tex
\usepackage{booktabs}
\usepackage{multirow}
\usepackage{graphicx}
\usepackage{latexsym}
\usepackage{amsmath}

\usepackage[T1]{fontenc}
\usepackage[utf8]{inputenc}

\usepackage{microtype}

\usepackage{graphicx}

\usepackage{booktabs}
\usepackage{multirow}
\usepackage{graphicx}
\usepackage{xcolor}

\newcommand{\scorecell}[2]{%
  \shortstack[c]{$#1$\\[-1pt]
  {\scriptsize\color{black!55}$\pm #2$}}%
}

\newcommand{\bestcell}[2]{%
  \shortstack[c]{$\mathbf{#1}$\\[-1pt]
  {\scriptsize\color{black!55}$\pm #2$}}%
}

\newcommand{\methodcell}[1]{%
  \shortstack[l]{#1\\[-1pt]
  {\scriptsize\phantom{$\pm 0.0$}}}%
}

\usepackage{tcolorbox}
\tcbuselibrary{breakable}

\usepackage{amsmath,booktabs,tabularx}

\newtcolorbox{promptbox}[1]{
    breakable,
    width=\linewidth,
    colback=white,
    colframe=black!45,
    colbacktitle=black!5,
    coltitle=black,
    title={#1},
    fonttitle=\bfseries\normalsize,
    fontupper=\normalfont\normalsize,
    boxrule=0.4pt,
    arc=1mm,
    boxsep=0pt,
    left=4pt,
    right=4pt,
    top=4pt,
    bottom=4pt,
    before skip=6pt,
    after skip=6pt
}

\setcitestyle{authoryear,round,semicolon}
\renewcommand{\cite}{\citep}

%% file: abstract.tex
The rapid growth of microdramas has increased demand for high-quality script production. Long-form microdrama script generation with LLMs requires both diverse plot progression and long-range narrative coherence. Our observations reveal that candidate continuations increasingly concentrate on fewer narrative mechanisms as generation progresses, while conventional retrieval may reinforce trope reuse and overlook historical information essential to future plots.
To address these challenges, we propose DraMaze, a multi-agent pipeline that explores diverse causal paths through trope memory-guided evolutionary tree search and mutation. It further constructs a Backward Narrative Dependency Tree to organize, retain, and retrieve historical memories based on future narrative goals. We also introduce MiDramaBench, a bilingual benchmark for evaluating script logic and creative quality across different script lengths, together with causal-pattern metrics for narrative repetition and diversity.
Experiments show consistent improvements across three backbone LLMs. For 60-episode scripts of up to 60k words, DraMaze reduces causal-pattern repetition by 32.1\% and achieves a Logic score of 82.5, outperforming the best-performing baseline by 8.5 points.

%% file: sections/02.tex
\section{Introduction}
\label{sec:introduction}

Microdramas have rapidly gained popularity worldwide as a form of fast-paced, serialized entertainment. LLM-based script generation offers a promising approach to meeting the growing demand for efficient and scalable content creation. However, maintaining narrative coherence and diverse plot progression across dozens of episodes remains challenging, despite LLMs' ability to produce fluent, well-structured scripts. As generation progresses, models may reuse familiar plot patterns despite changes in characters and events. For example, relationships may repeatedly advance through misunderstandings and clarifications. 
In microdramas, which are marked by rapid plot progression and twists, such repetition is especially noticeable and may make suspense-building turns predictable, thus weakening audience engagement.
% Such repetition is especially noticeable in microdramas, which feature rapid progression and frequent twists, potentially making turns intended to create suspense predictable and weakening audience engagement. 

% Prior research shows that AI-assisted writing can improve evaluations of individual works while reducing diversity across works, and that model-generated stories repeatedly exhibit similar plot elements~\citep{Doshi_2024,Xu_2025}, which may reflect generation preferences~\citep{holtzman2020curiouscaseneuraltext} and the narrative constraints accumulating in long-form creation. We further observe that, in the raw generating baseline, narrative mechanism entropy generally declines in the second half of the script, with candidate continuations concentrating on fewer plot-progression mechanisms (Section~3.1). This suggests that direct continuation alone struggles to sustain narrative mechanism diversity.

Prior work reports improved individual quality but reduced diversity across AI-assisted works, alongside recurring plot elements in generated stories~\citep{Doshi_2024,Xu_2025}. This may reflect generation preferences~\citep{holtzman2020curiouscaseneuraltext} and accumulating narrative constraints. Our experiment also shows declining narrative mechanism entropy in the script's second half, with candidate continuations concentrating on fewer progression mechanisms (Section~\ref{subsec:obs1}).

To enhance long-form generation, recent methods incorporate hierarchical planning, history summarization, and memory retrieval. DOME combines dynamic outlining with a temporal knowledge graph, while StoryWriter uses multi-agent planning and dynamic history compression~\citep{wang2024generatinglongformstoryusing,xia2025storywritermultiagentframeworklong}. While these approaches provide relevant historical context through memory management, they do not necessarily mitigate narrative homogenization. Actually, studies of LLM agents suggest that similar experiences and semantically relevant memories may shape subsequent decisions by encouraging models to repeat past behaviors or fixate on prior reasoning patterns~\citep{xiong2025memorymanagementimpactsllm,wang2026memtrapbenchbenchmarkingcognitivetraps}. Mechanism entropy still declines in later episodes with dense retrieval or DOME-style graph memory (Section~\ref{subsec:obs1}). In our continuation case study, both methods retrieve only 25.93\% of key historical information, and their continuations repeat past causal patterns more often than those generated with human-curated memories (Section~\ref{subsec:obs2}). These results suggest a gap between semantic relevance and narrative value: retrieval related to the current situation or characters may reintroduce previously used solutions while missing prerequisites for subsequent writing, weakening cross-episode coherence.

% These results indicate that historical relevance does not necessarily imply narrative value: retrieving events related to the current situation or characters may reintroduce previously used solutions while missing prerequisites for subsequent character choices, plot twists, and foreshadowing payoffs, potentially weakening cross-episode causal coherence and the credibility of plot developments.

% These observations suggest that long-form script generation should preserve essential facts without defaulting to past plot strategies. Coherent alternatives require both a clear narrative direction and the historical prerequisites needed to pursue it. Future narrative goals therefore guide both the exploration of diverse trajectories and the retention and retrieval of historical information.

These observations highlight the need to explore diverse plot trajectories without repeating past progression strategies, while preserving essential historical information. Since the value of historical information depends on its contribution to future plot development, future narrative goals should guide memory retention and retrieval.
% These observations suggest that long-form script generation should preserve the facts and causal prerequisites needed for future developments without treating past progression strategies as default solutions. Keeping new plot trajectories coherent requires considering both where the narrative is heading and which earlier setups make that development possible. Future narrative goals therefore guide both the exploration of diverse trajectories and the retention and retrieval of historical information. 

Based on this insight, we propose a multi-agent microdrama script generation pipeline, which integrates future goals into plot planning and episode-by-episode writing. 
% During planning, the pipeline uses evolutionary tree search to explore alternative causal paths around key narrative anchors, with narrative pattern memory identifying repeated progression strategies. It then selects event sequences that balance coherence, novelty, and goal reachability. 
During planning, it uses evolutionary tree search to explore diverse causal paths toward key narrative anchors, leveraging trope memory to discourage repeated progression mechanisms and mutation to broaden the search space.
Before episodic writing, the pipeline constructs a Backward Narrative Dependency Tree by tracing intermediate goals and their prerequisites backward from the planned ending.  During writing, this tree guides the selection and organization of key events and details, clues, and foreshadowing that support future developments, and enables retrieval according to the current episode's goals. 
Meanwhile, continually updated world-state memory maintains consistent states of characters, relationships, and assets.
By combining diverse plot planning with historical retrieval guided by future needs, the pipeline supports varied, causally coherent progression toward long-term narrative goals. Our main contributions are:
\begin{itemize}
    % \item We quantify diversity decline using narrative mechanism entropy and causal pattern reuse, revealing limitations of historical retrieval in recalling key information and reducing repetition.

    \item We reveal declining narrative mechanism diversity in long-form generation and show that conventional semantic retrieval may reinforce trope reuse while missing historical information essential to future plot developments.

    \item We propose \textbf{DraMaze}, a multi-agent pipeline that integrates \textbf{evolutionary plot planning} with \textbf{future-aware narrative memory} and dynamic state tracking to generate diverse yet coherent microdrama scripts.

    \item We introduce \textbf{MiDramaBench} for evaluating script logic and creative quality, together with causal-pattern metrics for quantifying narrative repetition and diversity across script lengths. Cross-model experiments and ablations validate the pipeline.
\end{itemize}

%% file: sections/03.tex
\section{Related Work}

\subsection{Long-Form Creative Writing Frameworks}
\label{sec:creative_writing}

% Although widely applied to creative writing, LLMs still face challenges in long-form generation, which have motivated approaches such as structured prompting, hierarchical planning, and iterative generation.Re3~\cite{yang2022re3generatinglongerstories} employs recursive prompting and revision, while DOC~\cite{yang2023docimprovinglongstory} improves narrative coherence through hierarchical outlining and constrained generation. DOME~\cite{wang2024generatinglongformstoryusing} combines dynamic hierarchical outlining with temporal knowledge graph memory for adaptive and coherent story generation.

% In addition, researchers have explored collaborative and adaptive generation frameworks. StoryWriter~\cite{xia2025storywritermultiagentframeworklong} coordinates outlining, planning, and writing agents to organize cross-chapter events and dynamically compress narrative history, while CritiCS~\cite{bae2024collectivecriticscreativestory} leverages collaborative critics to enhance creativity. WriteHERE~\cite{xiong2025outliningheterogeneousrecursiveplanning} employs adaptive task decomposition to enable flexible long-form writing. Beyond story, works like HoLLMwood~\cite{chen2024hollmwoodunleashingcreativitylarge} extends LLM-based creative writing to scripts generation through role-playing and multi-agent collaboration.

LLMs have been widely explored for creative writing, yet maintaining narrative quality in long-form generation remains challenging. Re3~\cite{yang2022re3generatinglongerstories} uses recursive prompting, candidate reranking, and revision, while DOC~\cite{yang2023docimprovinglongstory} employs detailed hierarchical outlines and controlled generation to improve narrative coherence. DOME~\cite{wang2024generatinglongformstoryusing} integrates dynamic hierarchical outlining with temporal knowledge graph memory to maintain long-range narrative consistency. Other relevant studies further explore specialized agent collaboration and adaptive planning. StoryWriter~\cite{xia2025storywritermultiagentframeworklong} coordinates outlining, planning, and writing agents for cross-chapter event organization and dynamic history compression.  CritiCS~\cite{bae2024collectivecriticscreativestory} iteratively refines story plans and drafts through collective critic feedback, while WriteHERE~\cite{xiong2025outliningheterogeneousrecursiveplanning} recursively decomposes writing into retrieval, reasoning, and composition tasks for adaptive generation. HoLLMwood~\cite{chen2024hollmwoodunleashingcreativitylarge} employs writer–editor–actor collaboration and role-playing in script generation. However, narrative repetition  and homogenization persist in long-form generation~\cite{chen2024hollmwoodunleashingcreativitylarge,Doshi_2024}, despite advances in plot planning and coherence.

% Nevertheless, prior studies have identified plot repetition and homogenization in LLM-generated creative content~\cite{chen2024hollmwoodunleashingcreativitylarge,Doshi_2024}. Despite advances in narrative coherence and writing quality, mitigating recurring narrative patterns while sustaining diverse plot developments throughout long-horizon generation remains an open challenge.

% Despite advances in narrative planning and coherence, plot repetition and homogenization remain challenges~\cite{chen2024hollmwoodunleashingcreativitylarge,Doshi_2024}, particularly in sustaining narrative diversity throughout long-form generation.

\subsection{Memory-Augmented Generation and Retrieval}
\label{sec:memory_retrieval}

% External memory and retrieval enable LLMs to leverage historical information for context-aware generation. Generative Agents~\cite{park2023generativeagentsinteractivesimulacra} integrates memory, reflection, and planning to support long-term behavior, while A-MEM~\cite{xu2025amemagenticmemoryllm} enables adaptive memory management through dynamic linking and update.

% In retrieval-augmented generation, semantic similarity-based retrieval is widely used to obtain relevant context. RAG~\cite{lewis2021retrievalaugmentedgenerationknowledgeintensivenlp} uses dense representations to ground generation in external knowledge, while RETRO~\cite{borgeaud2022improvinglanguagemodelsretrieving} incorporates similar text chunks through cross-attention to improve language modeling. REPLUG~\cite{shi2023replugretrievalaugmentedblackboxlanguage} refines document selection using language model feedback, whereas Adaptive-k~\cite{taguchi2025efficientcontextselectionlongcontext} dynamically adjusts the number of retrieved documents based on similarity scores. Recent studies have also explored the diversity and complementarity of retrieved information. SetR~\cite{lee2025shiftingrankingsetselection} improves coverage through set-level document selection, while DIVA~\cite{in2025diversifyverifyadaptefficientrobustretrievalaugmented} captures multiple interpretations of ambiguous queries through diversified retrieval.

Maintaining coherence in long-form generation requires effective organization and retrieval of narrative history. RecurrentGPT~\citep{zhou2023recurrentgptinteractivegenerationarbitrarily} maintains short- and long-term memories with semantic retrieval. SCORE~\citep{yi2025scorestorycoherenceretrieval} combines hierarchical summaries and state tracking with lexical-semantic retrieval, whereas Open-Theatre~\citep{xu2025opentheatreopensourcetoolkitllmbased} employs hierarchical memory and relevance-, importance-, and recency-based retrieval. Some studies also explore structured narrative memory. STORYTELLER~\citep{li2025storytellerenhancedplotplanningframework} builds an entity knowledge graph from subject–verb–object events for entity-based historical retrieval, while DOME~\cite{wang2024generatinglongformstoryusing} employs temporal knowledge graph memory with entity-based retrieval and semantic filtering. Narrative World Model \citep{saifullah2026narrativeworldmodelnarratologygrounded} represents evolving story states through a narratology-grounded temporal graph for multi-hop retrieval, while NstAgent \citep{wan2026scalinglongformstorygeneration} incrementally tracks character states, completed events, and unresolved narrative commitments. NarraWorld \citep{tian2026sharedworldsprivateminds} organizes world facts, character beliefs, ongoing developments, and possible continuations into interconnected memory views for situation-conditioned retrieval. While these methods further strengthen narrative consistency, organizing memories around the causal prerequisites of future narrative goals remains underexplored. 

% Despite these advances, effectively preserving, organizing, and retrieving historical events of future narrative value remains an open challenge.

%% file: sections/04.tex
\section{Empirical Analysis of Narrative Diversity and Retrieval}

% \subsection{Declining Diversity of Narrative Mechanisms}

\subsection{Declining Narrative Mechanism Entropy}

\label{subsec:obs1}

% As a script unfolds, LLMs may increasingly rely on similar mechanisms to advance the plot, despite changes in characters, settings, and events. This tendency persists, and may intensify, with semantic retrieval, which recalls past events resembling the current situation or involving key characters in the current episode. Such memories can encourage reuse of narrative devices and causal progression patterns. We quantify this tendency through the entropy of candidate narrative mechanisms.

As scripts progress, LLMs may increasingly rely on similar mechanisms to advance the plot, despite changes in characters, settings, and events. Semantic retrieval may reinforce this tendency by recalling past events resembling the current situation or involving the same characters, encouraging reuse of familiar narrative devices and causal progression patterns. We quantify this concentration through the entropy of candidate narrative mechanisms.

Using Qwen3.6-27B, we compare three strategies: full-history generation without retrieval (BASE), dense semantic retrieval of past events (Dense RAG), and DOME-style graph retrieval with semantic filtering (DOME). Each generates 40 episodes from the same premise. At generation checkpoints, we fix the context and retrieval results, then sample 32 candidate next events, each describing a main action and its immediate consequence.

% \begin{figure}[t]
%     \centering
%     \includegraphics[width=0.85\linewidth]{acl-style-files-master/figures/entropy2.pdf}
%     \caption{Narrative mechanism entropy across generation checkpoints.}
%     \label{fig:mechanism_entropy}
% \end{figure}
% \begin{figure}[t]
%     \centering
%     \includegraphics[width=0.85\linewidth]{acl-style-files-master/figures/memory_three_slot_repetition.png}
%     \caption{Narrative mechanism entropy across generation checkpoints.}
%     \label{memory_three_slot_repetition}
% \end{figure}

\begin{figure}[t]
    \centering
    \begin{minipage}[t]{0.495\textwidth}
        \vspace{0pt}
        \centering
        \includegraphics[width=0.85\linewidth]{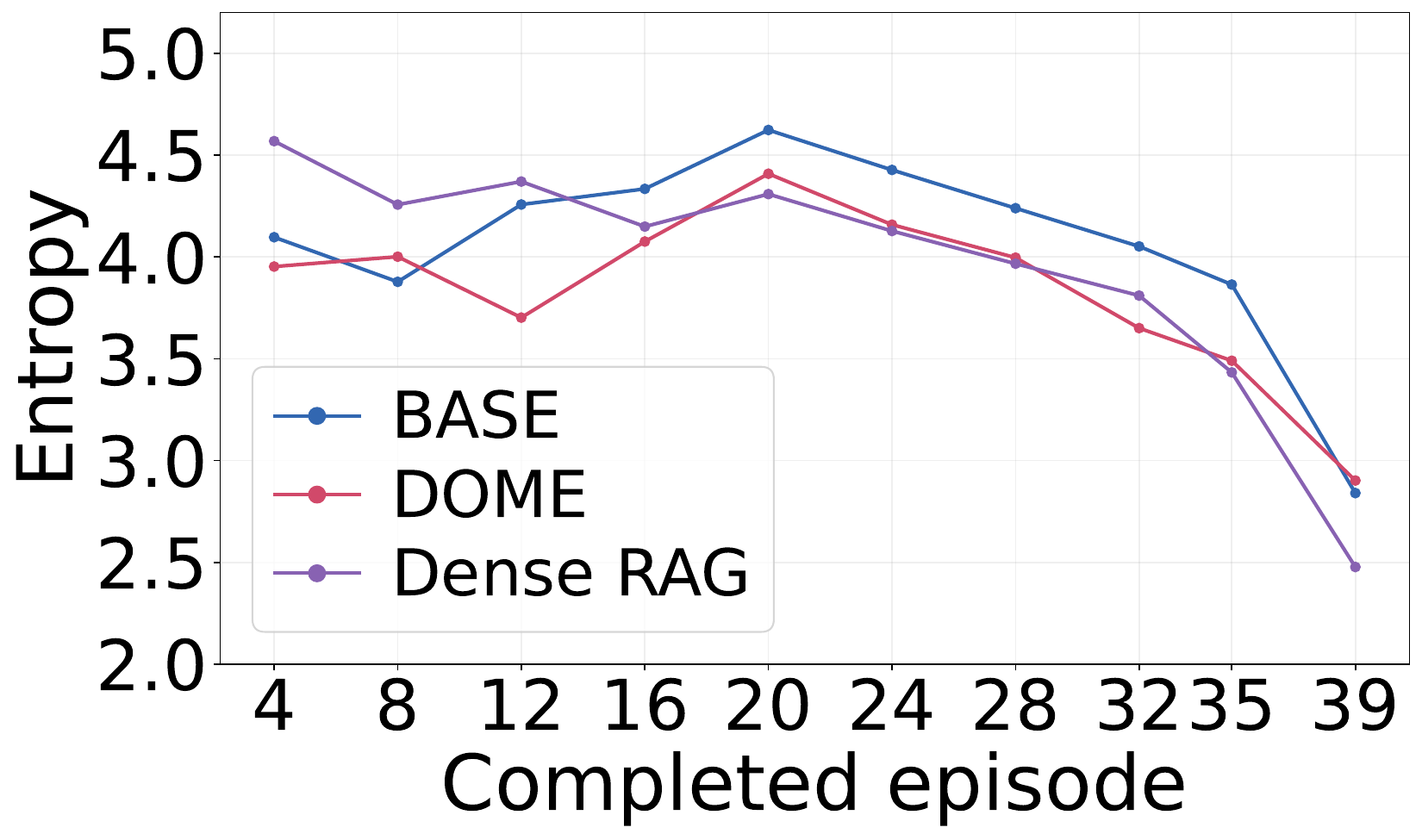}
        \caption{Narrative mechanism entropy across generation checkpoints.}
        \label{fig:mechanism_entropy}
    \end{minipage}\hspace{0.01\textwidth}%
    \begin{minipage}[t]{0.369\textwidth}
        \vspace{0pt}
        \centering
        \includegraphics[width=0.85\linewidth]{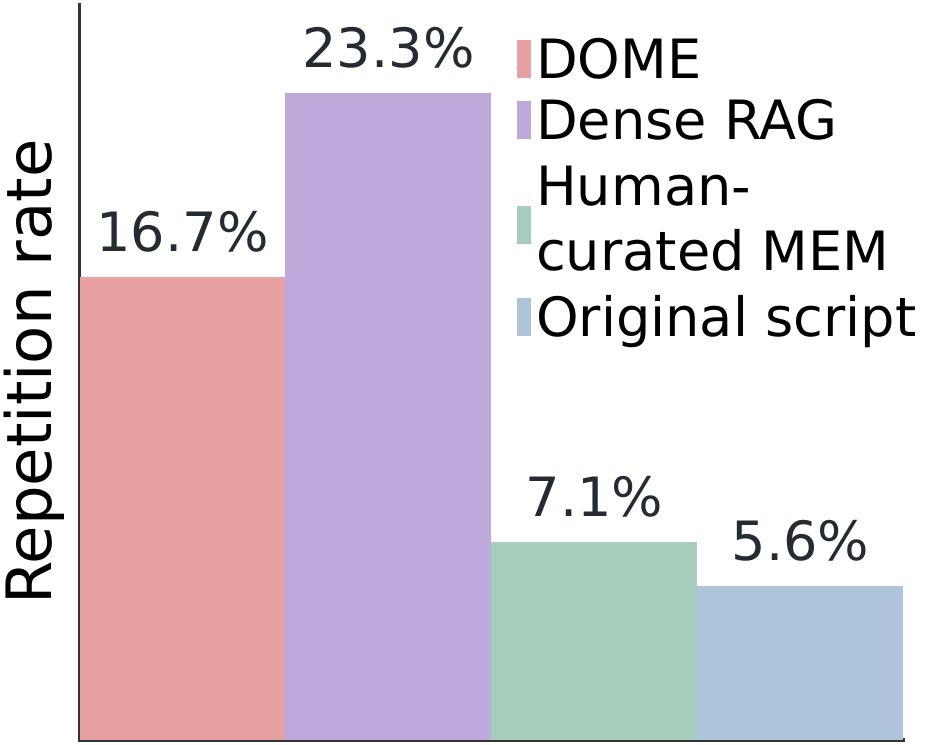}
        \caption{Historical trope repetition rates in different retrieval settings.}
        \label{fig:trope_repetition}
    \end{minipage}
\end{figure}

We define an event's \emph{core progression mechanism} as the means and causal process behind its first meaningful change to the situation. Qwen3.8-27B judges whether each pair of candidate events shares similar mechanisms. Deterministic greedy complete-link clustering then groups the candidates into mechanism clusters. For $N$ parseable events in $K$ clusters, narrative mechanism entropy is
\begin{equation}
H=-\sum_{k=1}^{K}p_k\log_2 p_k,
\qquad
p_k=\frac{|C_k|}{N},
\end{equation}
% \begingroup
% \small
% \begin{equation}
% H=-\sum_{k=1}^{K}p_k\log_2 p_k,
% \qquad
% p_k=\frac{|C_k|}{N},
% \end{equation}
% \endgroup
where $C_k$ is the $k$-th cluster. Lower entropy indicates greater concentration on fewer mechanisms. At each checkpoint, we report the mean entropy across five
scripts, with equal weight assigned to each script.

% Figure~\ref{fig:mechanism_entropy} shows that narrative mechanism entropy generally declines after episode 20, with a sharper drop near the end, indicating that candidate continuations increasingly concentrate on fewer narrative mechanisms. Notably, Dense RAG exhibits lower entropy than BASE throughout the second half, while DOME also falls below BASE at most later checkpoints. These results suggest that conventional retrieval methods fail to prevent this concentration and may even reinforce it, highlighting the need for more diverse yet coherent plot exploration.

Figure~\ref{fig:mechanism_entropy} shows that narrative mechanism entropy generally declines after episode 20, especially near the end. Dense RAG remains below BASE throughout the second half, while DOME does so at most later checkpoints. These results suggest that conventional retrieval fails to prevent, and may reinforce, concentration on fewer narrative mechanisms, highlighting the need for more diverse yet coherent plot exploration.

\subsection{The Semantic–Narrative Relevance Gap and Trope Reuse}
\label{subsec:obs2}
We further investigate whether semantic retrieval can identify historical information essential to future plot development and whether retrieved memories encourage reuse of existing narrative patterns.

We select a complete human-written microdrama script, using Episodes 1–20 as history and the synopses of Episodes 21–30 as future narrative goals. Dense RAG and DOME-style knowledge graph retrieval each retrieve the top 30 historical memories. We manually identify 27 historical subevents with narrative value for the continuation as references. Both methods retrieve only 7 of these 27 subevents, yielding a recall of \textbf{25.93\%}. This suggests that semantic or entity-level relevance does not necessarily reflect the causal dependencies between historical events and future narrative goals.

We further generate Episodes 21–30 under identical settings using memories retrieved by each method or human-curated memories, with the original human-written script as a reference. We represent narrative tropes as Cause–Method–Outcome and Method–Outcome–Response triples and measure their reuse from the historical episodes in the continuations. As shown in Figure~\ref{fig:trope_repetition}, average repetition rates are 23.30\% for Dense RAG and 16.67\% for DOME, compared with 7.14\% for human-curated memories and 5.56\% in the original script. These results suggest that automatic retrieval may not only overlook historical information essential to future developments but also encourage the reuse of existing conflict-resolution strategies and causal mechanisms.

Together, these observations reveal the persistent reuse of narrative mechanisms and a mismatch between semantic relevance and narrative function: semantically similar events may contribute little to current narrative goals, while essential causal prerequisites may be overlooked due to semantic differences. This motivates a shift from surface-level semantic matching toward retrieval guided by causal dependencies on future narrative goals, preserving necessary plot prerequisites while reducing unnecessary reuse of existing narrative patterns.

%% file: sections/05.tex
\section{Method}
\label{sec:method}

\subsection{Pipeline Overview}
\label{sec:pipeline-overview}
\begin{figure*}[t]
    \centering
    \includegraphics[width=0.98\textwidth]{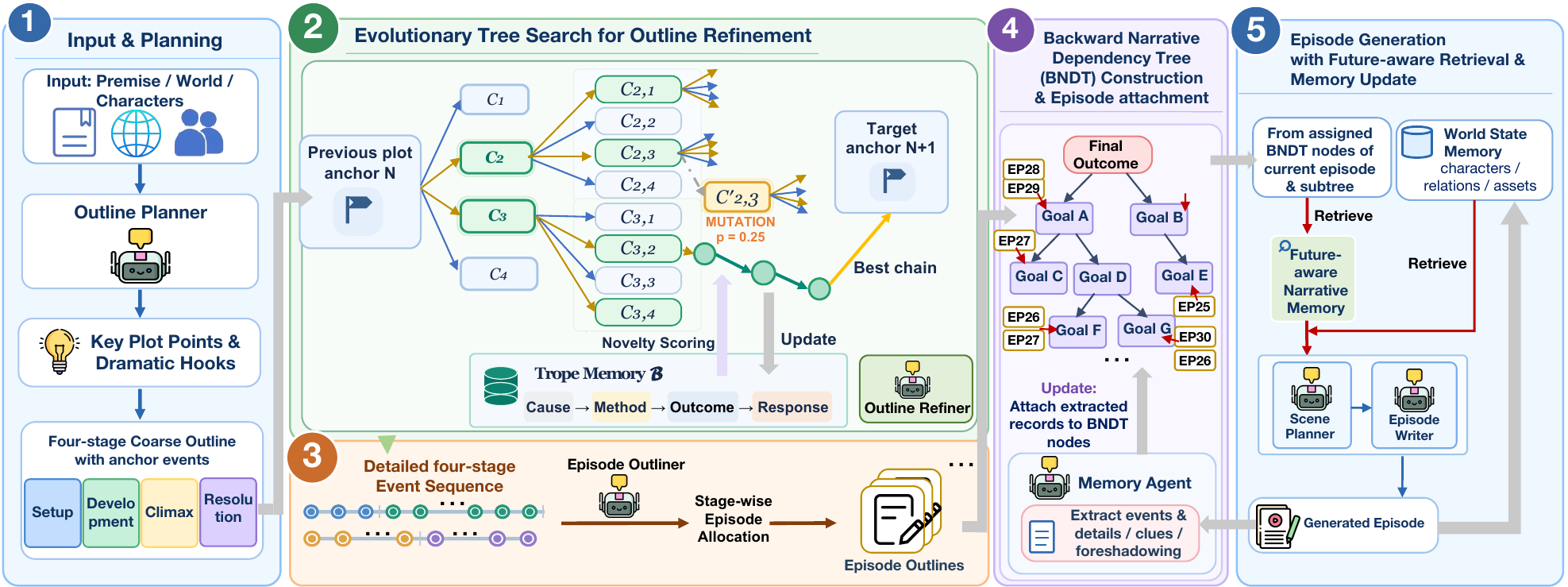}
    \caption{Overview of our multi-agent pipeline for microdrama script generation. (1) Outline Planner creates a four-stage coarse outline. (2) Evolutionary tree search refines event sequences through mutation and trope-guided selection. (3) Develop into episode outlines.(4) Backward Narrative Dependency Tree is constructed from episode outlines to organize subsequent future-aware narrative memory, with episodes attached to goal nodes to guide retrieval. (5) Agents retrieve memories, plan scenes, generate scripts, and update memories after each episode. }
    \label{fig:Overview}
\end{figure*}
As shown in Figure~\ref{fig:Overview}, our multi-agent pipeline develops an idea into a microdrama script through
hierarchical planning and sequential writing. Given the premise, world
setting, and characters, planning agents design key plot points and
dramatic hooks, then build a coarse outline spanning \emph{Setup},
\emph{Development}, \emph{Climax}, and \emph{Resolution}. Evolutionary
tree search expands its anchors into a detailed event sequence
(Section~\ref{sec:trope-search}), which is allocated across episodes and
developed into episode outlines. A memory-planning agent then works
backward from the ending to construct a \emph{Backward Narrative
Dependency Tree} (BNDT) for future-aware memory
(Section~\ref{sec:future-memory}). For each episode, agents retrieve narrative memories, plan scenes, write the script, and update memory
from the completed episode. Both mechanisms thus connect diverse narrative planning with consistent narrative execution.

\subsection{Evolutionary Tree Search for Outline Refinement}
\label{sec:trope-search}

The coarse outline specifies where the story goes, while refinement determines how it gets there. To avoid reusing solutions across different conflicts, we explore causal developments and explicitly evaluate their novelty against previously used mechanisms.

\paragraph{Trope Memory.}
A \emph{trope memory} $\mathcal{B}$ stores causal patterns extracted from
committed event sequences as \emph{cause--method--outcome--response}.
During extraction, we remove specific character names, locations, and
props to recognize recurring mechanisms despite surface differences. For example, bribing a guard with cash or jewelry uses
the same mechanism to gain cooperation. Patterns accumulate across all
four drama stages and subsequent candidates are compared against this
memory to score novelty.

\paragraph{Evolutionary Tree Expansion.}
For each narrative anchor in the coarse outline, we grow a
local search tree from the preceding plot. Each expanded node generates
four candidate next-event nodes. We score their novelty by comparing
the local causal tropes they form with preceding events against
trope memory $\mathcal{B}$, then select the top two for further expansion based on novelty and local narrative coherence.

Inspired by evolutionary algorithms, we introduce mutation with
probability $p_{\mathrm{mut}}=0.25$ at each expanded node. It
explicitly restructures the causal links in the highest-scoring
candidate's local trope while preserving the target anchor. The mutated event forms an additional branch to escape existing
causal patterns and broaden the search space.

Branches terminate upon reaching the anchor or search limit. Among
terminal chains $\mathcal{C}$, we select
\begin{equation}
  \begin{aligned}
    S(c)&=0.4C(c)+0.3N(c)+0.3F(c),\\
    c^*&=\operatorname*{arg\,max}_{c\in\mathcal{C}} S(c),
  \end{aligned}
  \label{eq:chain-selection}
\end{equation}
where $C$, $N$, and $F$ measure narrative consistency, trope novelty
relative to the accumulated trope memory $\mathcal{B}$, and the
feasibility of reaching the targeted anchor, respectively. The final
selected chain extends the plot and updates $\mathcal{B}$ with its causal
patterns to guide subsequent search.

\subsection{Future-Aware Narrative Memory}
\label{sec:future-memory}

\begin{figure*}[t]
    \centering
    \includegraphics[width=0.98\textwidth]{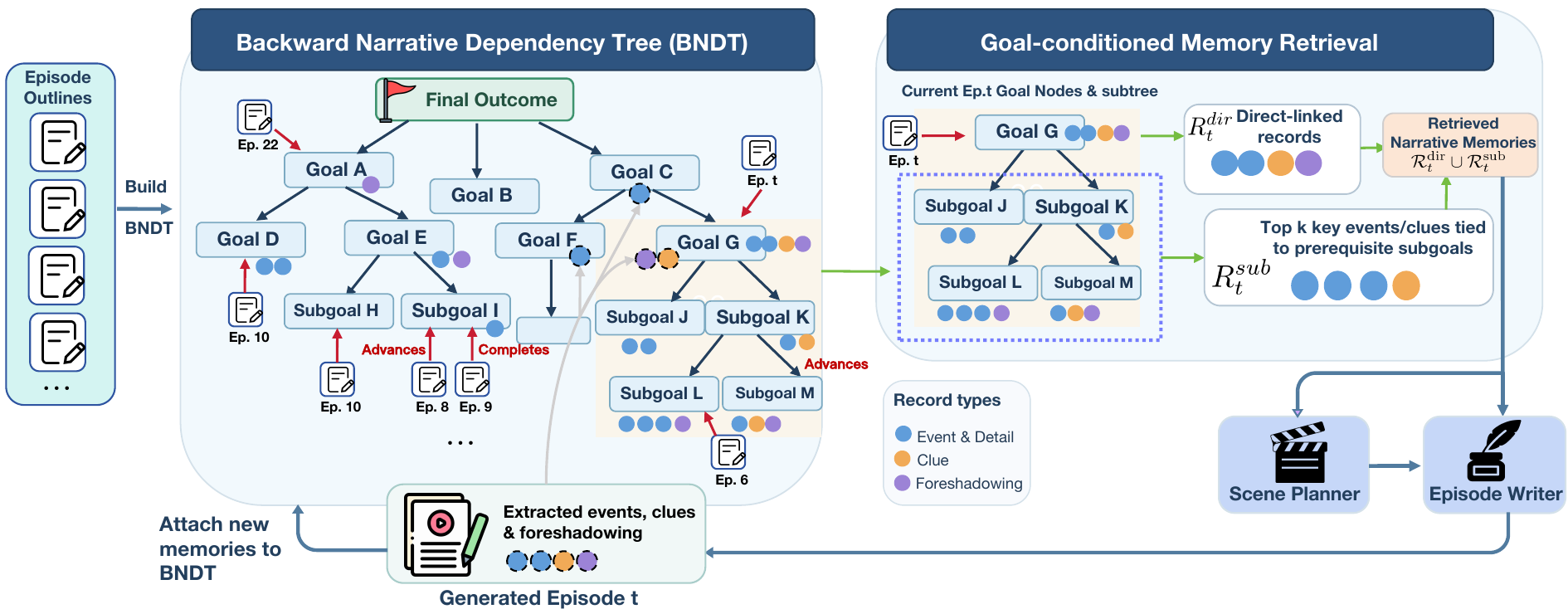}
    \caption{Overview of the Backward Narrative Dependency Tree (BNDT) and future-aware narrative memory retrieval. Episode outlines guide the backward construction of goal dependencies and episode attachment. Generated events, clues, and foreshadowing are linked to goal nodes, enabling retrieval from the current episode's goals and their prerequisite subtrees to support subsequent script generation.}
    \label{fig:BNDT}
\end{figure*}

The value of historical information depends on its contribution to future plot development, so we use future narrative goals to guide memory retention and retrieval. Figure~\ref{fig:BNDT} illustrates this process.

\paragraph{Backward Narrative Dependency Tree.}
The planning agent first extracts the final outcome from the episode
outlines and uses it as the root of the BNDT $\mathcal{F}=(V,E)$.
Working backward and guided by the episode outlines, it recursively
decomposes this outcome into intermediate goals and further
foundational setups. Each child specifies a
prerequisite for its parent goal. Then, each episode is
mapped to the goals $Q_t\subseteq V$ that it sets up, advances, or
completes. The tree provides a goal-based index for organizing and retrieving
narrative memories during script generation. An example can be seen in Appendix~\ref{casestudy}.

\paragraph{Narrative Memory Construction.}
After each episode is written, a memory agent extracts, guided by the established
BNDT, critical \emph{events \& details}, \emph{clues} with
long-term relevance, and \emph{foreshadowing} requiring future payoff.
It retains only information with future narrative
value for BNDT goals and links each memory record to the
corresponding goal nodes, recording its contribution and long-term
narrative importance score.
These links enable retrieval by narrative goal during subsequent
episode generation.

\paragraph{Narrative Memory Retrieval.}
Before writing episode $t$, we retrieve historical memories through its
assigned BNDT goal nodes $Q_t$. First, we collect all prior records
linked to at least one of $Q_t$:
\begin{equation}
  \mathcal{R}^{\mathrm{dir}}_t
  =\{m\in\mathcal{M}_{<t}:\Lambda(m)\cap Q_t\neq\emptyset\}.
  \label{eq:direct-memory}
\end{equation}
Here, $\mathcal{M}_{<t}$ contains memories from prior episodes, and
$\Lambda(m)$ denotes the BNDT goal nodes supported by record $m$.
The intersection selects memories sharing a narrative goal with the
current episode.

Next, we traverse subtrees beneath these goal nodes, as a few key events tied to
prerequisite subgoals may retain narrative value for current episode's
goals. From events
and clues linked to these descendant nodes, we retrieve up to
ten with the highest long-term narrative importance scores to form
$\mathcal{R}^{\mathrm{sub}}_t$, excluding already retrieved records.
Foreshadowing is retrieved only through
direct goal links, as it should have paid off at completed
prerequisite nodes. Their union,
$\mathcal{R}_t=\mathcal{R}^{\mathrm{dir}}_t\cup\mathcal{R}^{\mathrm{sub}}_t$,
provides narrative memories for the current goals.

\paragraph{World-State Memory.}
World-state memory maintains the current states of characters,
inter-character relationships, and assets. Before each episode is
generated, its outline is used to retrieve relevant state records through
textual relevance and entity matching.
These records are provided alongside narrative memories as context for
scene planning and script generation. After generation, changes
established in the completed script are used to update the stored
states, supporting consistent character and asset states across episodes.

%% file: sections/06.tex
\section{Evaluation}

% \begin{figure*}[t]
%     \centering
%     \includegraphics[width=0.8\textwidth]{acl-style-files-master/figures/evaluator-process.pdf}
%     \caption{Overview of the multi-agent evaluation framework for scripts logic and creative quality. Evidence-based review, auditing, and optional arbitration produce category-level ledger scores, which are combined with independent holistic scores using equal weights and averaged within each dimension.}
%     \label{fig:evaluator}
% \end{figure*}

\begin{figure}[t]
    \centering
    \includegraphics[width=0.90\columnwidth]{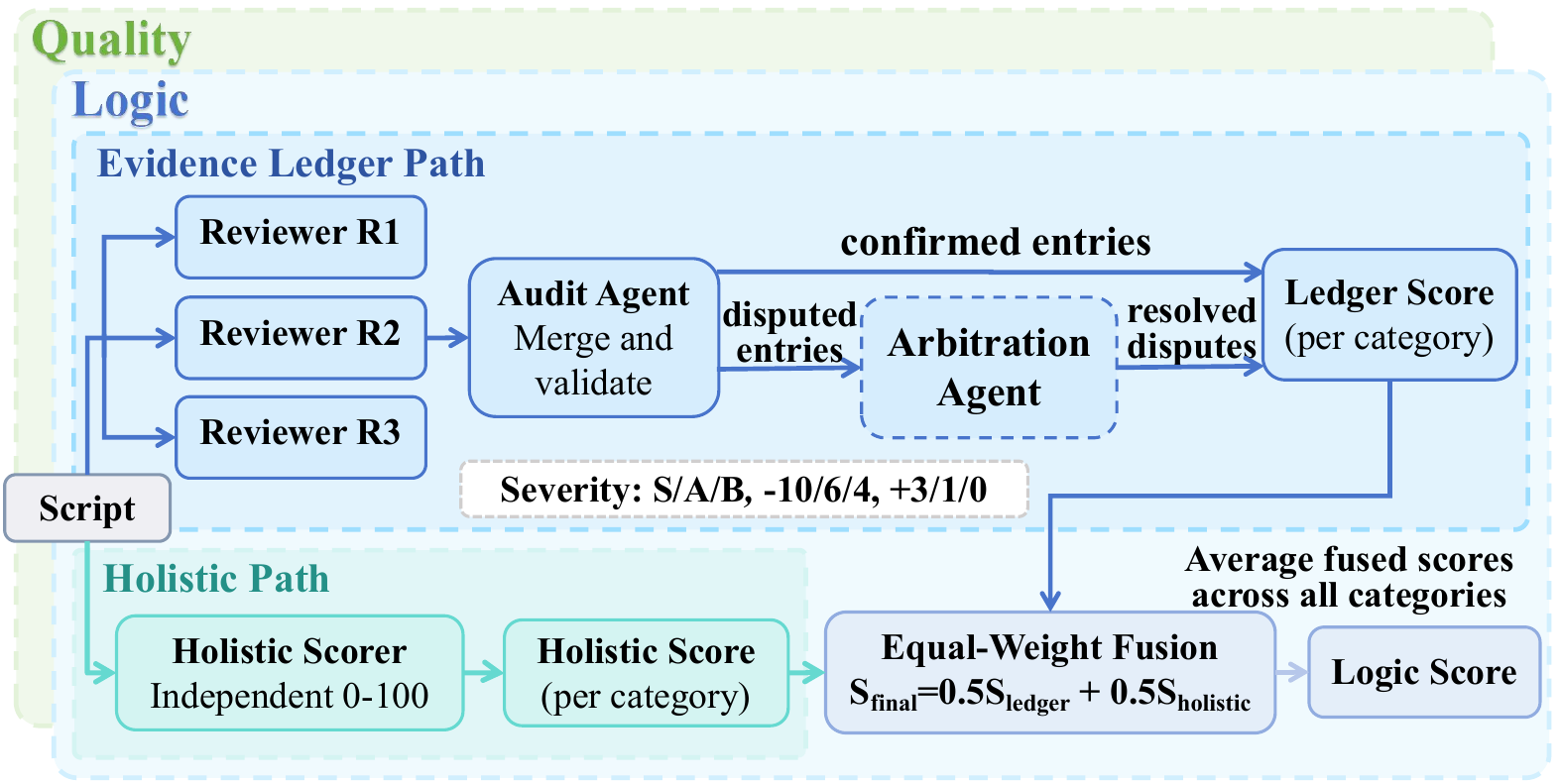}
    \caption{Overview of the multi-agent evaluation framework for script logic and creative quality.}
    \label{fig:evaluator}
\end{figure}

\subsection{Benchmark Construction}
\label{sec:benchmark}

\begin{figure}[htbp]
    \centering
    \includegraphics[width=0.98\textwidth]{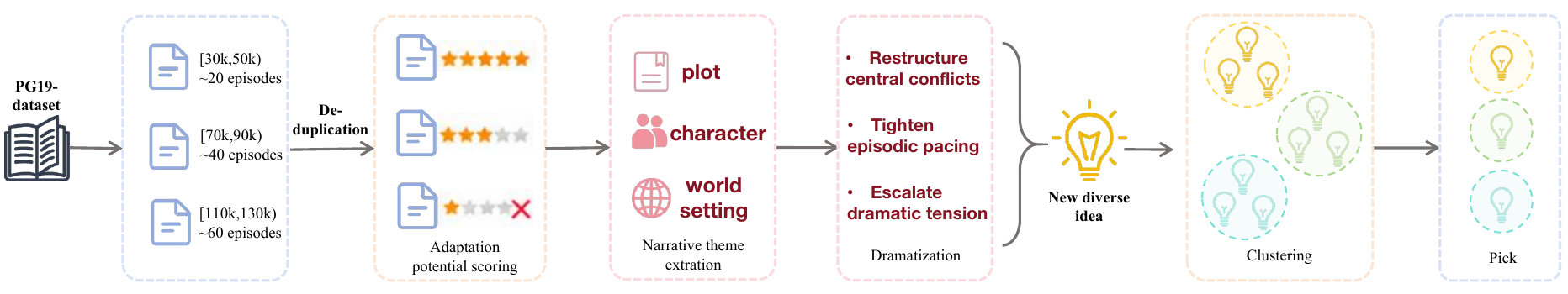}
    \caption{Construction pipeline of MiDramaBench. Source texts are
    selected by length, deduplicated, and screened for dramatic
    adaptation potential. Core plot developments, character relationships, and world settings are extracted to guide the dramatic adaptation of each source into a microdrama idea. Quality-reviewed ideas
    are grouped by genre-label embeddings and sampled within each
    length--cluster group to form the evaluation subset.}
    \label{fig:data}
\end{figure}

We construct a bilingual microdrama generation benchmark from PG-19
\citep{rae2019compressivetransformerslongrangesequence}, a public dataset of Project Gutenberg fiction and nonfiction, as illustrated in Figure~\ref{fig:data}. We select
sources in three length ranges, $[30k,50k)$, $[70k,90k)$, and
$[110k,130k)$ words, targeting 20-, 40-, and 60-episode dramas,
respectively, and deduplicate by content hashes and normalized titles.
These ranges target 700--1,000 English words per
episode, allowing for the condensation of source narration and
description during script adaptation. Then, we use Qwen3.8-27B to score
each source's dramatic adaptation potential based on narrative
continuity, character goals, actionable conflict, and episodic
expandability. For retained sources, the model extracts core
plot developments, character relationships, and world settings from
the full texts, then adapts them into open-ended
microdrama ideas by restructuring the central conflict for tightly paced
episodes with escalating tension, while preserving core
narrative. It then
scores the generated ideas on causal coherence,
dramatizability, episodic expandability, and creative openness,
retaining those that pass quality review after any necessary
revision.

For genre stratification, we encode accepted ideas' genre labels with BGE-large-en-v1.5 and apply k-means clustering to obtain
ten groups. We provide both English and Chinese versions of all ideas.
Within each length--cluster group, we retain up to
30 ideas ranked by review scores, yielding 899 candidates overall. We also construct
a stratified evaluation subset by sampling three distinct ideas from the top 20 per group: two for
English generation and one for Chinese generation. This yields 30 inputs
per length setting, with balanced
genre-cluster coverage within each language. 

\subsection{Logic and Quality Evaluator}

% To systematically assess the narrative logic and creative quality of long-form scriptss, we develop an LLM-based
% multi-agent evaluation framework by drawing on prior story evaluation criteria\citep{li2026loststoriesconsistencybugs,Yang_2025} and incorporating scripts-specific considerations\citep{ma2025dramabenchsixdimensionalevaluationframework}.
% The framework comprises two primary dimensions:

We develop an LLM-based multi-agent framework for evaluating microdrama scripts, with criteria co-designed with drama professionals and informed by narrative consistency error taxonomies~\citep{li2026loststoriesconsistencybugs}, reader-derived story quality criteria~\citep{Yang_2025}, and drama-specific evaluation dimensions~\citep{ma2025dramabenchsixdimensionalevaluationframework}. It assesses two dimensions:

\begin{itemize}
    \item \textbf{Logic:} causal chain, character state, knowledge state, resource state, rules and space-time, setup and payoff, repetition detection.
    \item \textbf{Quality:} character quality, dramatic engagement, plot-planning quality, episodic script quality, setting quality.
\end{itemize}

Each category is further divided into fine-grained criteria
for identifying narrative flaws and strengths. For example, within the character quality category,
the \emph{insufficient character arc} criterion identifies
protagonist development that lacks clear turning points
or consequential choices. Evaluation criteria
and detailed rules are provided in Appendix~\ref{app:evaluator}.

% We adopt a multi-stage evaluation process based on an Evidence Ledger. For each primary dimension, three independent LLM reviewers first identify narrative issues and strengths according to the predefined fine-grained criteria, recording supporting textual evidence along with the corresponding episode and scene locations. An Audit Agent then cross-checks the reviewers' findings, merges duplicate entries, and calibrates severity levels. An Arbitration Agent is further employed to resolve entries with remaining disagreements. Each confirmed entry is assigned one of three severity levels (S, A, or B). We employ a scoring scheme initialized at 90 points for each evaluation category. Identified issues incur deductions of 10, 6, or 4 points for severity levels S, A, and B, respectively, while identified strengths contribute bonuses of 3, 1, or 0 points. These adjustments yield an evidence-based score for each category, referred to as the Ledger Score.

As shown in Figure~\ref{fig:evaluator}, three independent LLM reviewers assess each dimension using predefined criteria, recording issues and strengths with textual evidence and episode locations in an Evidence Ledger. An Audit Agent validates and deduplicates entries and calibrates severity, while an Arbitration Agent resolves remaining disputes. Each category starts at 90 points, with S/A/B-level issues deducting 10/6/4 points and strengths adding 3/1/0 points, yielding the Ledger Score.

% We first adopt a multi-stage evaluation process based on an Evidence Ledger. For each primary dimension, three independent LLM reviewers identify issues and strengths using the predefined fine-grained criteria, recording supporting textual evidence along with the corresponding episode locations. An Audit Agent verifies findings, merges duplicates, and calibrates severity levels, while an Arbitration Agent further resolves remaining disagreements. Each category starts at 90 points, with confirmed issues deducting 10/6/4 points and strengths adding 3/1/0 points according to S/A/B severity levels, yielding the Ledger Score.

% To balance fine-grained error detection with holistic assessments of narrative quality, we further introduce an independent Holistic Scorer. The Holistic Scorer directly evaluates each category on a 0–100 scale, based on the complete scripts and the same evaluation criteria, without access to the previously constructed evidence ledger. The two scores are combined with equal weights: $S_{\mathrm{final}} = 0.5 S_{\mathrm{ledger}} + 0.5 S_{\mathrm{holistic}}.$

% The final Logic Score and Quality Score are then obtained by averaging the fused scores acros their respective categories. This design combines the interpretability of evidence-based evaluation with the comprehensive perspective of holistic quality assessment. The process can be seen in Fig~\ref{fig:evaluator}

To balance fine-grained error detection with holistic assessments of narrative quality, we further introduce an independent Holistic Scorer which evaluates each category on a 0--100 scale using the same criteria. Category scores are combined as $S_{\mathrm{final}} = 0.5 S_{\mathrm{ledger}} + 0.5 S_{\mathrm{holistic}}$, then averaged within each dimension to obtain the final Logic and Quality Scores. This design combines the interpretability of evidence-based evaluation with the broad perspective of holistic assessment.

\begin{table*}[t]
  \centering
  \small
  \setlength{\tabcolsep}{1.5pt}
  \renewcommand{\arraystretch}{1.15}

% ===== Table =====
\resizebox{\textwidth}{!}{%
\begin{tabular}{@{}cl*{17}{c}@{}}
    \toprule
    & & \multicolumn{8}{c}{\textbf{Script Logic}}
    & \multicolumn{6}{c}{\textbf{Script Quality}}
    & \multicolumn{3}{c}{\textbf{Homogeneity}} \\
    \cmidrule(lr){3-10}
    \cmidrule(lr){11-16}
    \cmidrule(lr){17-19}

    \textbf{Ep.} & \textbf{Method}
    & \textbf{CC} & \textbf{CS} & \textbf{KS}
    & \textbf{RS} & \textbf{RST} & \textbf{SP}
    & \textbf{RD} & \textbf{Avg}
    & \textbf{CQ} & \textbf{DE} & \textbf{PP}
    & \textbf{ES} & \textbf{SQ} & \textbf{Avg}
    & \textbf{4G}$\downarrow$
    & \textbf{3G}$\downarrow$
    & $\overline{H}\uparrow$ \\

    \midrule

    % ===== 20 Episodes =====
    \multirow{3}{*}{20}
    & \methodcell{DOME}
    & \scorecell{78.4}{1.3}
    & \scorecell{75.1}{0.8}
    & \scorecell{81.4}{0.3}
    & \scorecell{77.2}{0.7}
    & \scorecell{77.7}{0.4}
    & \scorecell{81.2}{1.0}
    & \scorecell{77.8}{1.2}
    & \scorecell{78.4}{0.1}
    & \scorecell{77.6}{0.4}
    & \scorecell{79.3}{1.0}
    & \scorecell{71.3}{0.9}
    & \scorecell{76.0}{1.0}
    & \scorecell{85.1}{0.9}
    & \scorecell{77.9}{0.5}
    & \scorecell{4.23}{0.60}
    & \scorecell{11.0}{0.15}
    & \scorecell{0.876}{0.0057} \\[1.3pt]

    & \methodcell{StoryWriter}
    & \scorecell{74.7}{0.3}
    & \scorecell{74.5}{0.7}
    & \scorecell{76.2}{0.4}
    & \scorecell{72.8}{1.0}
    & \scorecell{72.6}{1.1}
    & \scorecell{77.9}{0.4}
    & \scorecell{83.5}{0.5}
    & \scorecell{76.0}{0.3}
    & \scorecell{83.5}{0.8}
    & \scorecell{84.0}{0.3}
    & \scorecell{77.3}{1.0}
    & \scorecell{82.2}{0.2}
    & \scorecell{86.8}{0.1}
    & \scorecell{82.8}{0.1}
    & \scorecell{4.34}{0.64}
    & \scorecell{10.0}{1.1}
    & \scorecell{0.881}{0.0095} \\[1.3pt]

    & \methodcell{Ours}
    & \scorecell{81.5}{0.2}
    & \scorecell{80.4}{1.0}
    & \scorecell{82.6}{0.4}
    & \scorecell{80.1}{0.3}
    & \scorecell{81.3}{0.4}
    & \scorecell{83.7}{1.2}
    & \scorecell{88.5}{0.5}
    & \bestcell{82.6}{0.2}
    & \scorecell{87.3}{0.7}
    & \scorecell{87.7}{0.6}
    & \scorecell{82.9}{0.3}
    & \scorecell{85.2}{0.7}
    & \scorecell{87.6}{0.3}
    & \bestcell{86.1}{0.3}
    & \bestcell{1.35}{0.27}
    & \bestcell{5.51}{0.64}
    & \bestcell{0.939}{0.0054} \\

    \midrule

    % ===== 40 Episodes =====
    \multirow{3}{*}{40}
    & \methodcell{DOME}
    & \scorecell{73.0}{0.7}
    & \scorecell{65.3}{0.6}
    & \scorecell{77.6}{0.7}
    & \scorecell{73.0}{0.5}
    & \scorecell{71.1}{1.3}
    & \scorecell{76.1}{0.7}
    & \scorecell{68.8}{1.2}
    & \scorecell{72.1}{0.3}
    & \scorecell{77.0}{0.9}
    & \scorecell{78.4}{0.8}
    & \scorecell{69.4}{0.1}
    & \scorecell{76.3}{0.5}
    & \scorecell{85.1}{0.2}
    & \scorecell{77.3}{0.3}
    & \scorecell{7.86}{0.65}
    & \scorecell{17.3}{1.0}
    & \scorecell{0.883}{0.0034} \\[1.3pt]

    & \methodcell{StoryWriter}
    & \scorecell{73.5}{1.2}
    & \scorecell{70.8}{0.8}
    & \scorecell{76.3}{0.8}
    & \scorecell{72.3}{0.5}
    & \scorecell{71.2}{1.1}
    & \scorecell{77.4}{1.4}
    & \scorecell{81.1}{0.2}
    & \scorecell{74.6}{0.8}
    & \scorecell{85.9}{0.2}
    & \scorecell{84.3}{0.3}
    & \scorecell{78.1}{1.0}
    & \scorecell{82.0}{0.1}
    & \scorecell{87.4}{0.2}
    & \scorecell{83.5}{0.3}
    & \scorecell{6.65}{0.57}
    & \scorecell{15.2}{1.2}
    & \scorecell{0.894}{0.0025} \\[1.3pt]

    & \methodcell{Ours}
    & \scorecell{80.4}{0.6}
    & \scorecell{79.2}{0.3}
    & \scorecell{81.8}{0.5}
    & \scorecell{78.0}{1.2}
    & \scorecell{79.0}{0.6}
    & \scorecell{83.7}{0.7}
    & \scorecell{87.3}{0.9}
    & \bestcell{81.3}{0.7}
    & \scorecell{88.5}{0.3}
    & \scorecell{88.1}{0.6}
    & \scorecell{82.4}{0.8}
    & \scorecell{85.0}{0.4}
    & \scorecell{87.7}{0.1}
    & \bestcell{86.3}{0.4}
    & \bestcell{3.08}{0.41}
    & \bestcell{9.22}{0.63}
    & \bestcell{0.929}{0.0056} \\

    \midrule

    % ===== 60 Episodes =====
    \multirow{3}{*}{60}
    & \methodcell{DOME}
    & \scorecell{66.9}{0.2}
    & \scorecell{59.8}{0.4}
    & \scorecell{73.1}{0.4}
    & \scorecell{69.4}{1.1}
    & \scorecell{65.5}{0.3}
    & \scorecell{70.8}{0.4}
    & \scorecell{58.7}{1.1}
    & \scorecell{66.3}{0.1}
    & \scorecell{72.5}{1.3}
    & \scorecell{75.4}{0.6}
    & \scorecell{65.7}{1.0}
    & \scorecell{73.7}{0.4}
    & \scorecell{83.3}{0.5}
    & \scorecell{74.1}{0.1}
    & \scorecell{11.6}{0.52}
    & \scorecell{24.5}{0.72}
    & \scorecell{0.866}{0.0040} \\[1.3pt]

    & \methodcell{StoryWriter}
    & \scorecell{72.8}{1.1}
    & \scorecell{69.3}{1.2}
    & \scorecell{76.4}{0.7}
    & \scorecell{72.9}{1.3}
    & \scorecell{71.4}{1.3}
    & \scorecell{77.0}{0.7}
    & \scorecell{78.1}{0.3}
    & \scorecell{74.0}{0.8}
    & \scorecell{84.4}{1.4}
    & \scorecell{83.4}{0.6}
    & \scorecell{76.5}{0.6}
    & \scorecell{81.3}{0.5}
    & \scorecell{86.6}{0.2}
    & \scorecell{82.4}{0.4}
    & \scorecell{9.75}{0.90}
    & \scorecell{21.5}{0.91}
    & \scorecell{0.876}{0.011} \\[1.3pt]

    & \methodcell{Ours}
    & \scorecell{80.0}{0.1}
    & \scorecell{81.1}{0.6}
    & \scorecell{81.8}{0.5}
    & \scorecell{78.5}{0.1}
    & \scorecell{82.0}{0.6}
    & \scorecell{86.0}{0.7}
    & \scorecell{87.9}{0.9}
    & \bestcell{82.5}{0.2}
    & \scorecell{83.5}{0.9}
    & \scorecell{88.5}{0.3}
    & \scorecell{86.1}{0.3}
    & \scorecell{81.7}{0.5}
    & \scorecell{87.4}{0.2}
    & \bestcell{85.4}{0.1}
    & \bestcell{6.07}{0.32}
    & \bestcell{14.6}{0.64}
    & \bestcell{0.912}{0.0058} \\

    \bottomrule
    \end{tabular}
    }
    \caption{
    Main results on bilingual long-form script generation
    at 20, 40, and 60 episodes.
    Logic and Quality scores range from 0 to 100 (higher is better).
    \textbf{Logic:} CC (causal chain), CS (character state),
    KS (knowledge state), RS (resource state),
    RST (rules and space-time), SP (setup and payoff),
    and RD (repetition detection).
    \textbf{Quality:} CQ (character quality),
    DE (dramatic engagement), PP (plot-planning quality),
    ES (episodic script quality), and SQ (setting quality).
    Avg denotes the overall score for each dimension.
    \textbf{Homogeneity:} 4G denotes four-slot causal-pattern
    reuse (\%), 3G the average of two three-slot reuse rates
    (\%), and $\overline{H}$ the mean of three normalized
    conditional entropies (0--1).
    Cells report the mean (top) and sample standard deviation
    across three group means (bottom).
    Bold indicates the best Avg and Homogeneity scores
    within each episode group.
    }

  \label{tab:main-results}
\end{table*}

\subsection{Narrative Homogeneity Metrics}

% To evaluate narrative homogeneity in long-form scriptss, we develop a trope-based evaluator that measures the repetition and diversity of underlying narrative patterns rather than surface-level textual similarity. Specifically, we represent narrative patterns as four-slot causal chains consisting of \textbf{Cause, Method, Outcome, and Response (C--M--O--R)}, describing what triggers an event, how characters act, what consequences follow, and how characters subsequently respond. We first divide each scripts into overlapping episode windows and employ an LLM to extract causal chains, retaining their source episodes and plot-grounded causal explanations. Each slot is expressed as an abstract narrative mechanism, removing surface details such as character names and locations while preserving the underlying causal relationships.

To further assess narrative repetition and diversity, we develop causal-pattern metrics based on extracted and normalized causal chains. We represent narrative patterns as four-slot causal chains: \textbf{Cause, Method, Outcome, and Response (C--M--O--R)}, capturing event triggers, character actions, consequences, and subsequent reactions. 
% We first divide each scripts into overlapping episode windows and employ an LLM to extract causal chains, retaining source episodes and plot-grounded causal explanations. Each slot abstracts away character names and locations while preserving the underlying causal mechanisms.
An LLM extracts these chains from overlapping episode windows, abstracting away character names and locations while retaining causal mechanisms, source episodes, and plot-grounded explanations.

% Since semantically equivalent narrative mechanisms may be expressed differently across episodes, we perform \textbf{global semantic merging} to construct a unified taxonomy for each of the four slots, grouping expressions with equivalent narrative functions under the same category label. We then relabel the original causal chains by assigning the merged category labels to their corresponding slots, enabling consistent identification of narrative patterns across episodes. Further implementation details are provided in the Appendix\ref{}. Based on the relabeled causal chains, we compute three metrics to quantify narrative homogeneity.

% Since semantically equivalent narrative mechanisms may be expressed differently across episodes, we apply \textbf{global semantic merging} to construct a taxonomy for each of the four slots, grouping expressions with equivalent narrative functions. We then relabel the corresponding slots in the original causal chains using the merged category labels. Based on these relabeled chains, we compute three narrative homogeneity metrics. Further details are provided in Appendix~\ref{}.

Since the same underlying causal mechanisms may be expressed differently across episodes, we perform global causal-pattern canonicalization, grouping functionally equivalent expressions into a unified taxonomy for each slot and relabeling the extracted chains accordingly.  From these chains, we compute three metrics of narrative repetition and diversity.

\textbf{(1) 4-gram Repetition Rate.}
The proportion of C--M--O--R occurrences beyond each distinct pattern's first appearance.

% We measure the repetition of complete C--M--O--R patterns as the proportion of occurrences beyond the first appearance of each distinct pattern.

\textbf{(2) 3-gram Repetition Rate.}
The mean repetition rate of C--M--O and M--O--R triples.

% \textbf{(2) 3-gram Repetition Rate.}
% We measure partial pattern repetition by averaging the repetition rates of Cause--Method--Outcome (C--M--O) and Method--Outcome--Response (M--O--R) triples.

\textbf{(3) Conditional Entropy.}
Causal diversity is measured by support-weighted normalized conditional entropy:

% \begingroup
% \small
% \begin{equation}
% \begin{gathered}
% \widetilde{H}(Y\mid X)
% = \frac{1}{N_\tau}
% \sum_{x\in\mathcal{X}_\tau} n_x h_x,\\[3pt]
% h_x = \frac{H(Y\mid X=x)}
% {\log|\mathcal{Y}_x|}.
% \end{gathered}
% \end{equation}
% \endgroup
\begin{equation}
\begin{gathered}
\widetilde{H}(Y\mid X)
= \frac{1}{N_\tau}
\sum_{x\in\mathcal{X}_\tau} n_x h_x,\\[3pt]
h_x = \frac{H(Y\mid X=x)}
{\log|\mathcal{Y}_x|}.
\end{gathered}
\end{equation}

% where $n_x$ is the frequency of $x$, $\mathcal{X}_\tau=\{x:n_x\geq\tau\}$, $\mathcal{Y}_x$ denotes its observed successors, and $N_\tau=\sum_{x\in\mathcal{X}_\tau}n_x$. Single-successor cases contribute zero.

% The final entropy metric averages three causal transitions:
% $H_{\mathrm{avg}} = (\widetilde{H}(M\mid C) +
% \widetilde{H}(O\mid M) +
% \widetilde{H}(R\mid O))/3$.
% Higher repetition rates indicate greater pattern reuse,
% while lower entropy reflects reduced causal diversity.

Here, $n_x$ is the frequency of $x$,
$\mathcal{X}_\tau=\{x:n_x\geq\tau\}$,
$\mathcal{Y}_x$ is its observed successor set, and
$N_\tau=\sum_{x\in\mathcal{X}_\tau}n_x$.
% We set $h_x=0$ when $|\mathcal{Y}_x|=1$.
The final score is
$H_{\mathrm{avg}} = (\widetilde{H}(M\mid C) +
\widetilde{H}(O\mid M) +
\widetilde{H}(R\mid O))/3$.
Higher repetition and lower entropy indicate greater narrative homogeneity.

%% file: sections/07.tex
\section{Experimental Results}

% END GENERATED MAIN RESULTS

\subsection{Setup}

% We conduct experiments on the evaluation set of
% MiDramaBench, comprising 30 ideas (10 Chinese
% and 20 English) at 20, 40, and 60 episodes.
% For each method and length, we generate three
% scripts per idea. We report the mean and sample standard deviation across three groups, each with 30 ideas.

We evaluate on MiDramaBench's stratified evaluation set of 30 ideas (20 English, 10 Chinese) at 20, 40, and 60 episodes. All experiments are repeated three times, with results reported as mean and sample standard deviation across runs.

We use DOME and StoryWriter as baselines, adapting them to episodic script generation while
preserving their core workflows. DOME employs
five-stage hierarchical outlining with temporal
knowledge-graph memory, while StoryWriter uses
event decomposition, global episode planning,
writer--critic refinement, and history compression.

We use Qwen3.6-27B~\citep{qwen36_27b} as the backbone LLM for main generation experiments.
We further conduct experiments with GLM5.3-Flash~\citep{glm5team2026glm5vibecodingagentic} and DeepSeek-V4-Pro~\citep{deepseekai2026deepseekv4highlyefficientmilliontoken}
to assess the generalizability of our pipeline. For evaluation, we employ Qwen3.8-27B~\citep{qwen38} as the underlying model at temperature 0.1.

\begin{table}[!tbp]
\centering
\begin{minipage}[t]{0.48\textwidth}
\vspace{0pt}

  \centering
  \small
  \setlength{\tabcolsep}{2pt}
  \renewcommand{\arraystretch}{1.15}

  \scalebox{0.90}{%
  \begin{tabular}{@{}llccccc@{}}
    \toprule
    & & \textbf{Logic} & \textbf{Quality}
    & \multicolumn{3}{c}{\textbf{Homogeneity}} \\
    \cmidrule(lr){3-3}
    \cmidrule(lr){4-4}
    \cmidrule(lr){5-7}

    \textbf{Model} & \textbf{Method}
    & \textbf{Avg}$\uparrow$
    & \textbf{Avg}$\uparrow$
    & \textbf{4G}$\downarrow$
    & \textbf{3G}$\downarrow$
    & $\overline{H}\uparrow$ \\
    \midrule

    % ===== DeepSeek-V4-Pro =====
    \multirow{3}{*}{\shortstack[c]{DeepSeek-\\V4-Pro}}
    & \methodcell{DOME}
    & \scorecell{75.1}{0.4}
    & \scorecell{84.7}{0.5}
    & \scorecell{8.23}{0.70}
    & \scorecell{17.2}{0.43}
    & \scorecell{0.853}{0.012} \\[1.3pt]

    & \methodcell{StoryWriter}
    & \scorecell{73.3}{0.7}
    & \scorecell{84.7}{0.6}
    & \scorecell{6.74}{0.64}
    & \scorecell{14.5}{1.0}
    & \scorecell{0.829}{0.0068} \\[1.3pt]

    & \methodcell{Ours}
    & \bestcell{86.1}{0.2}
    & \bestcell{90.2}{0.2}
    & \bestcell{3.21}{0.57}
    & \bestcell{10.2}{0.51}
    & \bestcell{0.915}{0.0030} \\

    \midrule

    % ===== GLM5.3-Flash =====
    \multirow{3}{*}{\shortstack[c]{GLM5.3-\\Flash}}
    & \methodcell{DOME}
    & \scorecell{83.3}{0.4}
    & \scorecell{89.0}{0.3}
    & \scorecell{8.03}{0.94}
    & \scorecell{15.1}{0.94}
    & \scorecell{0.881}{0.0062} \\[1.3pt]

    & \methodcell{StoryWriter}
    & \scorecell{84.8}{0.4}
    & \scorecell{88.6}{0.2}
    & \scorecell{7.08}{1.1}
    & \scorecell{14.9}{0.90}
    & \scorecell{0.870}{0.010} \\[1.3pt]

    & \methodcell{Ours}
    & \bestcell{89.6}{0.2}
    & \bestcell{92.5}{0.0}
    & \bestcell{2.30}{0.19}
    & \bestcell{9.13}{0.77}
    & \bestcell{0.927}{0.0050} \\

    \bottomrule
  \end{tabular}
  }

  \caption{
    Cross-model generalization on 40-episode script generation using
    DeepSeek-V4-Pro and GLM5.3-Flash.
  }
  \label{tab:cross-model-results-average}

\end{minipage}\hfill
\begin{minipage}[t]{0.505\textwidth}
\vspace{0pt}

  \centering
  \small
  \setlength{\tabcolsep}{2pt}
  \renewcommand{\arraystretch}{1.15}

  \scalebox{0.90}{%
  \begin{tabular}{@{}lccccc@{}}
    \toprule
    & \textbf{Logic} & \textbf{Quality}
    & \multicolumn{3}{c}{\textbf{Homogeneity}} \\
    \cmidrule(lr){2-2}
    \cmidrule(lr){3-3}
    \cmidrule(lr){4-6}

    \textbf{Method}
    & \textbf{Avg}$\uparrow$
    & \textbf{Avg}$\uparrow$
    & \textbf{4G}$\downarrow$
    & \textbf{3G}$\downarrow$
    & $\overline{H}\uparrow$ \\
    \midrule

    % ===== Ours =====
    \methodcell{Ours}
    & \bestcell{81.3}{0.7}
    & \bestcell{86.3}{0.4}
    & \bestcell{3.08}{0.41}
    & \bestcell{9.22}{0.63}
    & \bestcell{0.929}{0.0056} \\[1.3pt]

    % ===== w/o Future-Aware Memory =====
    \methodcell{w/o Future-Aware Mem}
    & \scorecell{74.5}{0.3}
    & \scorecell{81.2}{0.3}
    & \scorecell{4.65}{0.41}
    & \scorecell{12.2}{0.86}
    & \scorecell{0.914}{0.0084} \\[1.3pt]

    % ===== w/o Memory and Tree Search =====
    \shortstack[l]{w/o Future-Aware Mem\\
    \& Tree Search}
    & \scorecell{72.1}{0.3}
    & \scorecell{81.2}{0.4}
    & \scorecell{7.13}{1.2}
    & \scorecell{16.4}{0.81}
    & \scorecell{0.887}{0.0091} \\[1.3pt]

    % ===== w/o Agent Pipeline =====
    \shortstack[l]{w/o Multi-Agent Pipeline\\
    (Direct Generation)}
    & \scorecell{70.3}{0.1}
    & \scorecell{76.8}{0.2}
    & \scorecell{6.87}{0.14}
    & \scorecell{16.8}{0.68}
    & \scorecell{0.887}{0.012} \\

    \bottomrule
  \end{tabular}
  }

  \caption{
    Ablation results on 40-episode script
    generation using Qwen3.6-27B.
    We progressively remove Future-Aware
    Memory and Evolutionary Tree Search while
    retaining the original Agent Pipeline.
    Direct removes the Multi-Agent Pipeline entirely.
  }
  \label{tab:ablation-results}

\end{minipage}
\end{table}

\subsection{Main Results}

Table~\ref{tab:main-results} presents the results of
long-form script generation at 20, 40, and 60
episodes using Qwen3.6-27B. 
Our method
consistently outperforms DOME and StoryWriter in
script logic and quality across all episode lengths.
% While baseline Logic scores decline as generation length
% increases, Ours remains stable (82.6 at 20 episodes
% vs. 82.5 at 60), with its advantage over the strongest
% baseline widening from 4.2 to 8.5 points.
Across the three length settings, DraMaze achieves similar Logic scores (82.6, 81.3, and 82.5 for 20, 40, and 60 episodes, respectively), while its margin over the strongest baseline increases from 4.2 to 8.5 points.
The improvements extend across key aspects of
long-range narrative consistency. Meanwhile, the consistently high
Quality scores indicate that improved coherence is
accompanied by sustained creative quality.

% Table~\ref{tab:main-results} presents the results of
% long-form scripts generation at 20, 40, and 60
% episodes using Qwen3.6-27B. Our method consistently
% outperforms DOME and StoryWriter in both scripts
% logic and quality, with a growing advantage as
% the generation length increases. While the baselines
% exhibit declining logic scores over longer horizons,
% Ours maintains stable performance (82.6 at 20 episodes
% vs. 82.5 at 60), with its margin over the strongest
% baseline widening from 4.2 to 8.5 points.
% The improvements extend across key aspects of
% long-range narrative consistency, particularly
% character states, causal chains, and setup and payoff,
% suggesting better preservation of narrative dependencies
% across episodes. Meanwhile, the consistently high
% Quality scores indicate that improved coherence is
% accompanied by sustained creative quality rather than
% achieved at its expense.

Ours also achieves the lowest 4-gram and 3-gram
repetition rates and the highest conditional entropy
across all lengths. At 60 episodes, 3-gram repetition
drops to 14.6\%, compared with 21.5\% for the
strongest baseline. Although structural repetition increases with script
length, our method consistently maintains greater
narrative diversity, demonstrating its effectiveness
in mitigating narrative homogenization over extended
generation.

% Our method also consistently achieves lower narrative
% homogeneity. Across all episode lengths, Ours yields
% the lowest 4-gram and 3-gram repetition rates and the
% highest conditional entropy. At 60 episodes, the
% 3-gram repetition rate decreases to 14.6\%, compared
% with 21.5\% for the strongest baseline.
% Although structural repetition generally increases
% with scripts length, our method maintains a clear
% advantage in narrative diversity, indicating that
% improved long-range coherence does not come at the
% cost of repetitive plot patterns. 

Cross-model results in Table~\ref{tab:cross-model-results-average} and Table~\ref{tab:cross-model-results}
further demonstrate the generalizability of our method.
On DeepSeek-V4-Pro and GLM5.3-Flash, ours improves
Logic scores by 4.8--11.0 points and Quality scores by
3.5--5.5 points over the strongest baselines, while
consistently reducing repetition and increasing
conditional entropy. These results confirm its
effectiveness across different backbones.

% To examine cross-model generalization, we further
% evaluate our method on DeepSeek-V4-Pro and GLM5.3-Flash
% with 40-episode scriptss, as shown in
% Table~\ref{tab:cross-model-results}.
% On DeepSeek-V4-Pro, Ours achieves Logic and Quality
% scores of 86.1 and 90.2, outperforming the strongest
% baselines by 11.0 and 5.5 points, respectively.
% On GLM5.3-Flash, the corresponding scores reach
% 89.6 and 92.5, with gains of 4.8 and 3.5 points.
% Moreover, our method consistently achieves the
% lowest repetition rates and the highest conditional
% entropy on both models. These findings demonstrate
% that the improvements in narrative coherence,
% creative quality, and diversity generalize across
% different backbone LLMs.

\subsection{Ablation Study}

% END GENERATED ABLATION RESULTS

To assess component contributions, we conduct ablations on 40-episode scripts using Qwen3.6-27B. We progressively remove Future-Aware Memory and Evolutionary Tree Search while retaining the multi-agent pipeline. We also include a direct-generation baseline producing five episodes per iteration with full history as context.

As shown in Table~\ref{tab:ablation-results}, removing future-aware narrative memory reduces Logic/Quality from 81.3/86.3 to 74.5/81.2, increases 3-gram repetition from 9.22\% to 12.2\%, and lowers entropy from 0.929 to 0.914. These results support the contribution of future-aware narrative memory to coherence and reduced causal-pattern repetition within the full pipeline. Further removing tree search raises 4-gram repetition from 4.65\% to 7.13\% and lowers entropy to 0.887, highlighting its role in narrative diversity. Direct generation further reduces Logic/Quality to 70.3/76.8, which demonstrates the value of hierarchical planning and multi-agent coordination.

%% file: sections/08.tex
\section{Conclusion}
% This work investigates narrative homogenization in long-form scripts generation, revealing a gap between semantic relevance and future narrative needs. To address this gap, we propose a Future-aware Script Generation Pipeline that integrates trope-guided evolutionary tree search, narrative memory organized around future goals and their prerequisites, and dynamic world-state tracking. The accompanying MiDramaBench provides bilingual generation tasks, logic and quality evaluation, and causal-pattern metrics for narrative repetition and diversity. Experiments on 20, 40 and 60-episode scripts and across multiple backbone models show consistent improvements in logic, creative quality, and diversity, while ablations support the complementary contributions of future-aware memory and plot exploration. These findings highlight the value of aligning planning and memory selection with future narrative goals to sustain coherent and diverse long-form generation.
This work investigates narrative homogenization in long-form script generation, revealing a gap between semantic relevance and future narrative needs. To address these challenges, we propose DraMaze, combining trope-guided evolutionary tree search, narrative memory organized around future goals, and dynamic world-state tracking. We further introduce MiDramaBench for bilingual script generation and evaluation of logic and creative quality, together with causal-pattern metrics for narrative repetition and diversity. Experiments on 20-, 40-, and 60-episode scripts across multiple backbone LLMs demonstrate consistent improvements in logic, creative quality, and diversity, while ablations validate key component effectiveness. These findings highlight the value of combining diverse plot exploration with future-goal-guided memory retrieval for coherent and diverse long-form generation.

%% file: references.tex
\bibliography{custom}

%% file: appendices/01.tex
\appendix
\raggedbottom

%% file: appendices/02.tex
\section{Details of the Mechanism Entropy Experiment}
\label{app:mechanism_entropy}

\subsection{Checkpoint Sampling}
\label{app:entropy_sampling}

We probe each 40-episode script after episodes
$t\in\{4,8,12,16,20,24,28,32,35,39\}$.
At each checkpoint, we freeze the premise, episode index,
recent context and retrieval results,
then independently sample 32 next-event candidates per group.
Each candidate describes a character's main action and its direct consequence in approximately 80--150 tokens.
Candidates are not truly appended to the script or used to
condition subsequent samples. Both episode generation and event sampling use Qwen3.6-27B.

\subsection{Mechanism Similarity Evaluation}
\label{app:mechanism_judgment}

For each event, the evaluator independently identifies
the first substantive progression unit: the key means
that address the opening predicament or advance the
ongoing main action, together with the resulting change
in the situation. 

The comparison focuses on how events overcome or create obstacles, reveal or conceal information, shift character incentives, or alter control relationships. 
Differences in characters, locations, or objects are
incidental when they preserve the causal mechanism. 

Two events are judged similar when their identified progression
units share the same key causal means, and dissimilar when those
means clearly differ. Each pair is evaluated in both input orders
and treated as similar for clustering only when both judgments
agree on similarity. All within-group pairwise evaluations are
completed before clustering.

\subsection{Clustering and Entropy Computation}
\label{app:entropy_clustering}

\paragraph{Deterministic complete-link clustering.}
Let $s_{ij}=1$ if events $i$ and $j$ are stably similar,
and $s_{ij}=0$ otherwise.
Each parseable event initially forms a singleton cluster,
ordered by its original sample index.
Two clusters $C_a$ and $C_b$ can merge only when
\begin{equation}
s_{ij}=1
\quad
\text{for every } i\in C_a,\ j\in C_b.
\end{equation}
We scan cluster pairs in lexicographic order of their
positions and merge the first eligible pair.
After each merge, members are sorted by sample index,
clusters are reordered lexicographically by their
member-index sequences, and scanning restarts.
The procedure terminates when no eligible pair remains.
This rule ensures that all events within a cluster are
pairwise stably similar. For example, if pairs $(0,1)$
and $(0,2)$ are stably similar but $(1,2)$ is not,
the algorithm first merges events 0 and 1, yielding
$\{\{0,1\},\{2\}\}$.

\paragraph{Group-level entropy.}
For each sampling group, we compute entropy from the
cluster-size distribution using the definition in
Section~\ref{subsec:obs1}.
The denominator is the number of parseable events.
Groups are clustered and scored independently.

\paragraph{Aggregation across scripts.}
Let $H_{s,t}^{(m)}$ denote the entropy computed from the
32 next-event candidates sampled for script $s$ under
method $m$ at checkpoint $t$. We average these entropy
values across the five scripts:
\begin{equation}
\bar{H}_{t}^{(m)}
=
\frac{1}{S}
\sum_{s=1}^{S} H_{s,t}^{(m)},
\qquad S=5.
\end{equation}

% \paragraph{Aggregation across scripts.}
% Let $H_{s,t,g}^{(m)}$ denote the entropy of sampling group
% $g$ for script $s$, method $m$, and checkpoint $t$.
% We first average group entropies within each script,
% then average across the five scripts:
% \begin{equation}
% \bar{H}_{t}^{(m)}
% =
% \frac{1}{S}
% \sum_{s=1}^{S}
% \left(
% \frac{1}{G_{s,t}^{(m)}}
% \sum_{g=1}^{G_{s,t}^{(m)}} H_{s,t,g}^{(m)}
% \right),
% \end{equation}
% where $S=5$ and $G_{s,t}^{(m)}$ is the number of groups
% included for that script, method, and checkpoint.
% Each script therefore contributes equally to the
% reported curve.

%% file: appendices/03.tex
\section{Benchmark Data Example}
\label{app:benchmark-construction}

Table~\ref{tab:short-drama-idea-example} presents a representative generated microdrama idea from MiDramaBench. Each idea comprises four fields: genre, core plot, characters, and world setting, which collectively establish the central conflict, character motivations and relationships, and relevant setting constraints. As illustrated by the example, these structured inputs provide sufficient narrative grounding for long-form generation while leaving room for diverse drama plot developments. This allows different generation methods to explore and organize episode-level narratives from a common starting point.

\begin{table*}[t]
  \centering
  \small
  \setlength{\tabcolsep}{5pt}
  \renewcommand{\arraystretch}{1.18}
  \begin{tabular}{@{}p{0.15\textwidth}p{0.79\textwidth}@{}}
    \toprule
    \textbf{Field} & \textbf{Input idea} \\
    \midrule
    Genre & Corporate thriller; family drama. \\
    \addlinespace[2pt]
    Core plot & A former CEO, publicly ousted for a scandal he did not commit,
    returns to his family company disguised as a janitor. He must expose board
    members who are siphoning assets and forcing his estranged wife into a merger
    that would erase her legacy. When his son, a junior executive investigating
    his father's disappearance, becomes the board's next target, the search for
    evidence becomes a race to protect him and prevent the company's dissolution.
    The CEO's low-status disguise lets him move unnoticed through the hierarchy,
    gather evidence, and test loyalties while risking exposure at every encounter.
    \\
    \addlinespace[2pt]
    Characters & The former CEO is observant and strategically restrained, but
    secrecy distances him from his family. His impulsive, loyal son is desperate
    to establish his authority and sees the disguised father as a stranger. A
    politically adept board rival seeks control of the company and the son's
    removal. The CEO's wife is a resourceful executive who delays the merger
    while investigating the suspicious new janitor. \\
    \addlinespace[2pt]
    World Setting& A closed corporate environment in which reputation, hierarchy,
    and access to information determine power. Nondisclosure agreements and
    internal governance restrict outside legal recourse, making an internal
    resolution necessary. Access credentials and artifacts such as a keycard
    or private server can serve as proof of authority. \\
    \bottomrule
  \end{tabular}
  \caption{Example English microdrama idea with four fields: genre, core plot, characters, and world setting.}
  \label{tab:short-drama-idea-example}
\end{table*}

%% file: appendices/04.tex
\section{Case Study: BNDT Construction and Episode Attachment}
\label{casestudy}
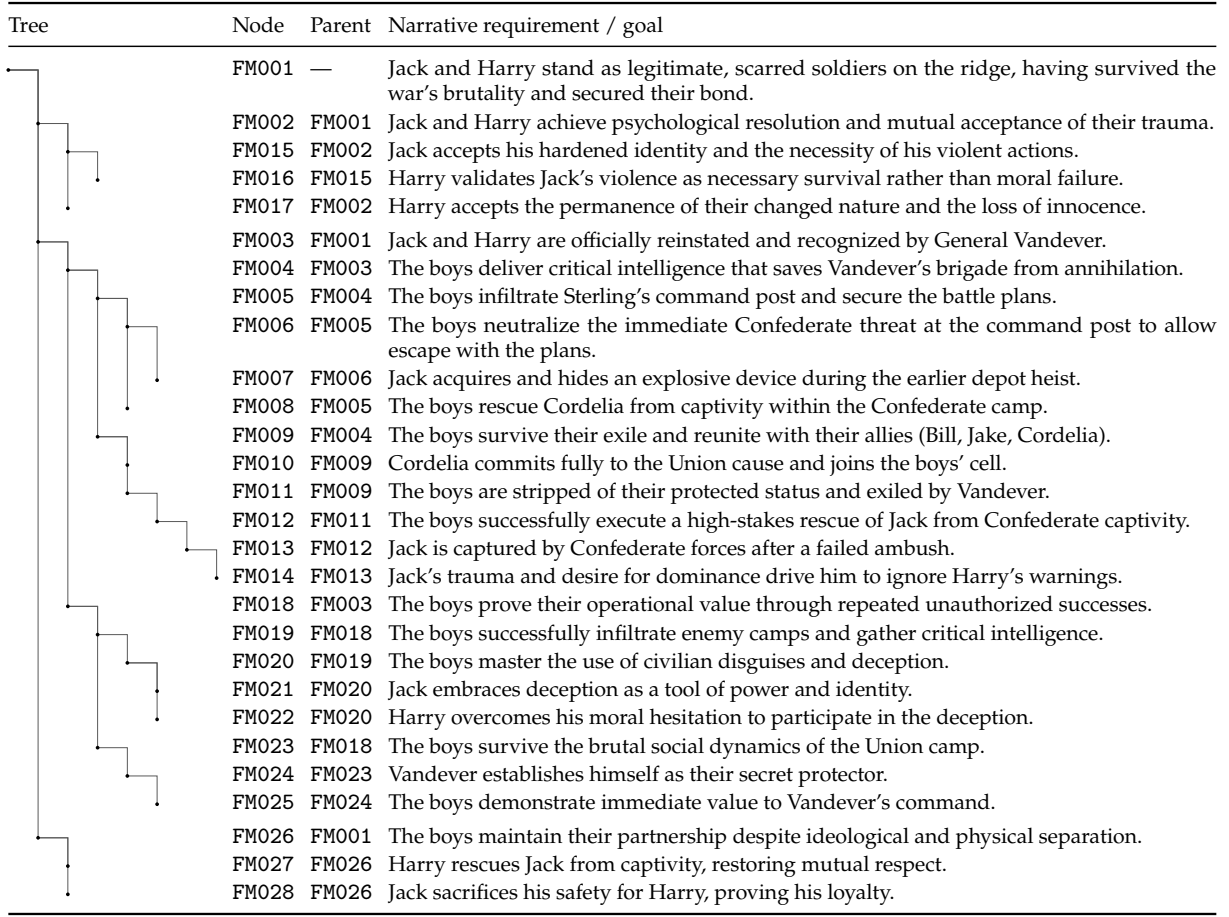
\begin{figure*}[t]
  \centering
  \begingroup
  \scriptsize
  \setlength{\tabcolsep}{3pt}
  \renewcommand{\arraystretch}{1.12}
  \newcommand{\bndtmark}[2]{%
    \hspace*{#1em}%
    \tikz[remember picture,baseline=-0.5ex]
      \coordinate (n#2);%
  }
  \begin{tabular}{@{}lllp{0.66\textwidth}@{}}
    \toprule
    Tree & Node & Parent & Narrative requirement / goal \\
    \midrule
    \bndtmark{0}{FM001} & \texttt{FM001} & --- & Jack and Harry stand as legitimate, scarred soldiers on the ridge, having survived the war's brutality and secured their bond. \\
    \bndtmark{1.4}{FM002} & \texttt{FM002} & \texttt{FM001} & Jack and Harry achieve psychological resolution and mutual acceptance of their trauma. \\
    \bndtmark{2.8}{FM015} & \texttt{FM015} & \texttt{FM002} & Jack accepts his hardened identity and the necessity of his violent actions. \\
    \bndtmark{4.2}{FM016} & \texttt{FM016} & \texttt{FM015} & Harry validates Jack's violence as necessary survival rather than moral failure. \\
    \bndtmark{2.8}{FM017} & \texttt{FM017} & \texttt{FM002} & Harry accepts the permanence of their changed nature and the loss of innocence. \\
    \addlinespace[2pt]
    \bndtmark{1.4}{FM003} & \texttt{FM003} & \texttt{FM001} & Jack and Harry are officially reinstated and recognized by General Vandever. \\
    \bndtmark{2.8}{FM004} & \texttt{FM004} & \texttt{FM003} & The boys deliver critical intelligence that saves Vandever's brigade from annihilation. \\
    \bndtmark{4.2}{FM005} & \texttt{FM005} & \texttt{FM004} & The boys infiltrate Sterling's command post and secure the battle plans. \\
    \bndtmark{5.6}{FM006} & \texttt{FM006} & \texttt{FM005} & The boys neutralize the immediate Confederate threat at the command post to allow escape with the plans. \\
    \bndtmark{7}{FM007} & \texttt{FM007} & \texttt{FM006} & Jack acquires and hides an explosive device during the earlier depot heist. \\
    \bndtmark{5.6}{FM008} & \texttt{FM008} & \texttt{FM005} & The boys rescue Cordelia from captivity within the Confederate camp. \\
    \bndtmark{4.2}{FM009} & \texttt{FM009} & \texttt{FM004} & The boys survive their exile and reunite with their allies (Bill, Jake, Cordelia). \\
    \bndtmark{5.6}{FM010} & \texttt{FM010} & \texttt{FM009} & Cordelia commits fully to the Union cause and joins the boys' cell. \\
    \bndtmark{5.6}{FM011} & \texttt{FM011} & \texttt{FM009} & The boys are stripped of their protected status and exiled by Vandever. \\
    \bndtmark{7}{FM012} & \texttt{FM012} & \texttt{FM011} & The boys successfully execute a high-stakes rescue of Jack from Confederate captivity. \\
    \bndtmark{8.4}{FM013} & \texttt{FM013} & \texttt{FM012} & Jack is captured by Confederate forces after a failed ambush. \\
    \bndtmark{9.8}{FM014} & \texttt{FM014} & \texttt{FM013} & Jack's trauma and desire for dominance drive him to ignore Harry's warnings. \\
    \bndtmark{2.8}{FM018} & \texttt{FM018} & \texttt{FM003} & The boys prove their operational value through repeated unauthorized successes. \\
    \bndtmark{4.2}{FM019} & \texttt{FM019} & \texttt{FM018} & The boys successfully infiltrate enemy camps and gather critical intelligence. \\
    \bndtmark{5.6}{FM020} & \texttt{FM020} & \texttt{FM019} & The boys master the use of civilian disguises and deception. \\
    \bndtmark{7}{FM021} & \texttt{FM021} & \texttt{FM020} & Jack embraces deception as a tool of power and identity. \\
    \bndtmark{7}{FM022} & \texttt{FM022} & \texttt{FM020} & Harry overcomes his moral hesitation to participate in the deception. \\
    \bndtmark{4.2}{FM023} & \texttt{FM023} & \texttt{FM018} & The boys survive the brutal social dynamics of the Union camp. \\
    \bndtmark{5.6}{FM024} & \texttt{FM024} & \texttt{FM023} & Vandever establishes himself as their secret protector. \\
    \bndtmark{7}{FM025} & \texttt{FM025} & \texttt{FM024} & The boys demonstrate immediate value to Vandever's command. \\
    \addlinespace[2pt]
    \bndtmark{1.4}{FM026} & \texttt{FM026} & \texttt{FM001} & The boys maintain their partnership despite ideological and physical separation. \\
    \bndtmark{2.8}{FM027} & \texttt{FM027} & \texttt{FM026} & Harry rescues Jack from captivity, restoring mutual respect. \\
    \bndtmark{2.8}{FM028} & \texttt{FM028} & \texttt{FM026} & Jack sacrifices his safety for Harry, proving his loyalty. \\
    \bottomrule
  \end{tabular}
  \begin{tikzpicture}[remember picture,overlay]
    \begin{scope}[draw=black!65,line width=0.35pt]
      \draw (nFM001) -| (nFM002);
      \draw (nFM001) -| (nFM003);
      \draw (nFM001) -| (nFM026);
      \draw (nFM002) -| (nFM015);
      \draw (nFM002) -| (nFM017);
      \draw (nFM015) -| (nFM016);
      \draw (nFM003) -| (nFM004);
      \draw (nFM003) -| (nFM018);
      \draw (nFM004) -| (nFM005);
      \draw (nFM004) -| (nFM009);
      \draw (nFM005) -| (nFM006);
      \draw (nFM005) -| (nFM008);
      \draw (nFM006) -| (nFM007);
      \draw (nFM009) -| (nFM010);
      \draw (nFM009) -| (nFM011);
      \draw (nFM011) -| (nFM012);
      \draw (nFM012) -| (nFM013);
      \draw (nFM013) -| (nFM014);
      \draw (nFM018) -| (nFM019);
      \draw (nFM018) -| (nFM023);
      \draw (nFM019) -| (nFM020);
      \draw (nFM020) -| (nFM021);
      \draw (nFM020) -| (nFM022);
      \draw (nFM023) -| (nFM024);
      \draw (nFM024) -| (nFM025);
      \draw (nFM026) -| (nFM027);
      \draw (nFM026) -| (nFM028);
    \end{scope}
    \foreach \id in {FM001,FM002,FM003,FM004,FM005,FM006,FM007,FM008,
      FM009,FM010,FM011,FM012,FM013,FM014,FM015,FM016,FM017,FM018,
      FM019,FM020,FM021,FM022,FM023,FM024,FM025,FM026,FM027,FM028}
      \fill (n\id) circle (0.7pt);
  \end{tikzpicture}
  \endgroup
  \caption{A complete backward narrative dependency tree (BNDT) generated for a 40-episode wartime drama. Orthogonal connectors show parent--child dependencies and each child is a prerequisite for its parent. All 28 node goals are reproduced from the generated tree, in depth-first order.}
  \label{fig:bndt-i17-r01}
\end{figure*}
We illustrate BNDT construction and episode attachment using a 40-episode wartime microdrama centered on Jack and Harry. Figure~\ref{fig:bndt-i17-r01} presents the complete dependency tree constructed backward from the final outcome, where each child goal represents a prerequisite for its parent. Episodes are attached to goal nodes according to whether they set up, advance, or complete the corresponding goals.

% For example, the outline for Episode 25 centers on Harry's covert rescue of Jack after Vandever refuses to help, rebuilding their trust. It is thus linked to FM027 (“Harry rescues Jack from captivity, restoring mutual respect”) as complete. 
For example, the outline for Episode 26 has Vandever declare the boys “dead to the army” and send them out of
camp with Bill and Jake. This completes FM011 (“The boys are stripped of their protected status and exiled by
Vandever”).
In contrast, Episode 32 outlines Jack impersonating an officer and bluffing past a sentry with a fabricated
order from Sterling. This advances FM005 (“The boys infiltrate Sterling’s command post and secure the battle plans”), since they gain entry, but have yet to obtain the plans.
An episode may also serve multiple goals. Episode 19 outlines “Jack kills a sentry to help his companions escape, but is shaken by his own violence.” It advances FM018 (“The boys prove their operational value through repeated unauthorized successes”) while setting up FM015 (“Jack accepts his hardened identity and the necessity of his violent actions”).

%% file: appendices/05.tex
\section{Details of Logic and Quality Evaluation}
\label{app:evaluator}

This section presents the evaluation criteria and representative
rating anchors that guide evidence identification, severity
assessment, and scoring in Logic and Quality evaluation.

\subsection{Evaluation Criteria}

Logic comprises seven scoring categories, while Quality comprises five.
Table~\ref{tab:evaluator-criteria} summarizes their principal checks.
Each category is further divided into fine-grained criteria that guide
the identification of narrative issues and strengths and specify
the evidence required for their assessment. All agents receive
the same dimension-specific rubric, including the applicable severity
anchors and scope boundaries.

\begin{table*}[t]
\centering
\small
\setlength{\tabcolsep}{4pt}
\renewcommand{\arraystretch}{1.12}
\begin{tabularx}{\textwidth}{@{}p{0.12\textwidth}p{0.21\textwidth}X@{}}
\toprule
Dimension & Category & Principal checks \\
\midrule
Logic & Causal chain & Prerequisite information, action motivation, causal support for key turns, and justification of crisis resolution. \\
& Character state & Continuity of life, identity, allegiance, abilities, personality, motivation, and relationships. \\
& Knowledge state & Information acquisition, knowledge sources, retention of established facts, and the scope of secret disclosure. \\
& Resource state & Ownership and transfer, depletion and restoration, evidence provenance, and support for resource increases. \\
& Rules and space-time & Established rules, preparation of new rules, temporary exceptions, event chronology, and spatial movement. \\
& Setup and payoff & Abandoned setups, forced payoffs, unprepared revelations, and missing repercussions of major events. \\
& Repetition detection & Replayed events and repeated episode boundaries that add no substantive development and impede progression. \\
\midrule
Quality & Character quality & Protagonist distinctiveness, arc and depth; supporting-character and antagonist functions; ensemble balance; emotional bonds, relational tension, and relationship progression. \\
& Dramatic engagement & Opening setup and protagonist introduction; clarity of goals, opposition, and stakes; sustained anticipation, cross-episode follow-through, payoff, and emotional investment. \\
& Plot-planning quality & Event non-repetition, obstacle differentiation, scene progression, opening and later pacing, emotional trajectory, continuing anticipation, information distribution, padding, subplot integration, and main-plot payoff. \\
& Episodic Script quality & Character-specific and situation-appropriate dialogue, emotional movement, information density, spoken naturalness, filmable description, scene staging, and concrete action. \\
& Setting quality & Genre fit, contribution of setting to the plot, exposition load, and pacing of setting revelation. \\
\bottomrule
\end{tabularx}
\caption{Summary of the Logic and Quality evaluation criteria. }
\label{tab:evaluator-criteria}
\end{table*}

\paragraph{Evidence and rating anchors.}
Each finding must identify a named check and cite supporting evidence with its episode and scene location. Contradictions and repetition require evidence from both relevant passages, while positive ratings must be supported by developments across episodes or stages, or by significant plot points. The severity of identified issues is then assessed using criterion-specific anchors, considering their scope, frequency, evidential clarity, and the extent of revision required.

\subsection{Detailed Rating Anchors for Character State}
\label{app:character-state-anchors}

We illustrate the fine-grained rubric using the Character State
category. Its six issue criteria are life-state discontinuity,
identity or allegiance discontinuity, contradictory character
identity, ability-state discontinuity, abrupt change of
personality or motivation, and abrupt relationship reversal.
Each finding must cite specific textual evidence with episode
and scene locations; contradictions require both relevant
passages. S/A/B denote issue severity. Its core content is presented below.

\begin{promptbox}{Life-State Discontinuity}
Assess whether later appearances and actions respect established
life states and physical constraints.

\par\smallskip
\noindent\textbf{S.}
A character explicitly and realistically confirmed dead later
participates as a living person in a key event, without an
explanation such as a staged death, resurrection, double,
or flashback.

\par\smallskip
\noindent\textbf{A.}
A severe constraint, such as grave injury, coma, or imprisonment,
disappears without treatment, escape, or an account of elapsed
time.

\par\smallskip
\noindent\textbf{B.}
Recovery is unusually rapid or thinly explained, but does not
invalidate the plot.
\end{promptbox}

\begin{promptbox}{Identity or Allegiance Discontinuity}
Assess whether identity revelations and changes in allegiance
are supported and reflected in subsequent events.

\par\smallskip
\noindent\textbf{S.}
An identity reveal or betrayal has occurred and been noticed,
yet later events proceed as if it had not happened, invalidating
a key plot development.

\par\smallskip
\noindent\textbf{A.}
A change in identity or allegiance lacks sufficient events or
emotional motivation, making cooperation or opposition
conspicuously abrupt.

\par\smallskip
\noindent\textbf{B.}
The change has a basis, but its emotional transition and
supporting scenes are weak.
\end{promptbox}

\begin{promptbox}{Contradictory Character Identity}
Assess whether established identity facts and relationship
labels remain consistent across the script.

\par\smallskip
\noindent\textbf{S.}
A core identity, family order, title, office, blood tie, or
marital relationship directly contradicts an earlier statement
and affects a main-plot premise.

\par\smallskip
\noindent\textbf{A.}
An important identity label, form of address, or relationship
label is inconsistent, hindering understanding of relationships,
organizational position, or authority.
\end{promptbox}

\begin{promptbox}{Ability-State Discontinuity}
Assess whether the use and development of abilities respect
previously established restrictions and changes.

\par\smallskip
\noindent\textbf{S.}
An ability explicitly removed, sealed, exhausted, or restricted
is later used normally to reverse a crisis without the
restriction being lifted.

\par\smallskip
\noindent\textbf{A.}
An upgrade, awakening, or removal of a restriction lacks the
necessary process, cost, or trigger.

\par\smallskip
\noindent\textbf{B.}
The change has a basic explanation, but its development is
thin or unusually rapid.
\end{promptbox}

\begin{promptbox}{Abrupt Change of Personality or Motivation}
Assess whether changes in established traits or goals are
supported by events and psychological development.

\par\smallskip
\noindent\textbf{S.}
A core personality trait or long-term goal reverses
fundamentally without a major event, psychological setup,
or emotional buildup.

\par\smallskip
\noindent\textbf{A.}
An event motivates the change, but its significance clearly
cannot support the magnitude of the reversal.

\par\smallskip
\noindent\textbf{B.}
A local action slightly conflicts with established
characterization without undermining overall consistency.
\end{promptbox}

\begin{promptbox}{Abrupt Relationship Reversal}
Assess whether major changes in interpersonal relationships
follow from sufficient events and emotional development.

\par\smallskip
\noindent\textbf{S.}
Hatred, trust, intimacy, cooperation, or betrayal reverses
fundamentally without a key event, emotional buildup,
or resolution of a misunderstanding.

\par\smallskip
\noindent\textbf{A.}
An event motivates the reversal, but its impact is insufficient
or the buildup is rushed.

\par\smallskip
\noindent\textbf{B.}
The shift is inferable, but the emotional transition or
its expression is thin.
\end{promptbox}

%% file: appendices/06.tex
\section{Details of Trope Homogeneity Evaluation}
\label{app:trope-evaluator}

The trope homogeneity evaluator measures repetition and diversity
in causal narrative patterns through causal-chain extraction,
normalization, and metric computation. Extracted
Cause--Method--Outcome--Response (C--M--O--R) chains are mapped
to unified category labels within each script, from which
repetition rates and conditional entropy are computed.

\subsection{Causal-Chain Extraction}

The reference configuration uses seven-episode windows with a two-episode overlap to capture cross-episode causal chains spanning window boundaries. An LLM extracts meaningful micro-level
causal chains across all subplots.

Each chain consists of four slots:
\textbf{Cause}, the event or state directly prompting an action;
\textbf{Method}, the action and its operative mechanism;
\textbf{Outcome}, the change directly produced by that action;
and \textbf{Response}, a character's subsequent reaction.

The extraction instructions require concise, reusable
phrases that preserve the mechanism and direction of change
while abstracting away names, locations, and specific props.
Each chain also includes traceable textual evidence from the original script and an explicit explanation of the corresponding causal relationships.
To avoid duplicate counting across overlapping windows, we deduplicate chains referring to the same underlying narrative event based on their episode locations and textual evidence.
% Chains preserve event order as presented in the text.
% The instructions exclude unsupported causal links, unrelated
% event combinations, unnecessary fragmentation, and recalled
% events treated as new occurrences. Overlapping episodes
% provide context: a chain is retained only if it includes
% an actual event in the newly covered episodes.
% A later reference to an earlier event alone does not satisfy
% this condition. Evidence episodes identify actual chain
% events within the current window.

\subsection{Normalization}

We construct a global taxonomy for each C--M--O--R slot using phrases extracted across all windows. Guided by plot-grounded causal explanations, we group expressions sharing the same core mechanism or behavioral goal, direction of change, and relevant subject--object relationships. A substitution test further verifies whether replacing one expression with another preserves its causal meaning within the slot. Thus, surface differences do not prevent merging, while shared themes or forms of expression alone are insufficient.

For example, bribing a guard with money to gain passage and offering jewelry to an informant for information both rely on exchanging benefits for cooperation and should be grouped together. In contrast, threatening a guard to gain passage relies on coercion rather than exchange, and should remain distinct despite achieving the same outcome.
% \paragraph{Consensus calibration.}
% Three independent taxonomy proposals are reconciled by
% a consensus calibrator using the original chains,
% their explanations, and phrase-coverage diagnostics.
% The calibrator evaluates disagreements in category boundaries
% and granularity against the same semantic criteria.
% It splits categories that conflate distinct mechanisms
% and merges categories separated only by interchangeable
% surface expressions. Agreement across proposals does not
% replace semantic validation.

% Calibration maintains a consistent degree of semantic
% differentiation across slots without requiring equal
% category counts. Singleton categories are retained for
% semantically distinct mechanisms.
% Each original phrase must appear verbatim in exactly one
% category within its own slot. Coverage diagnostics identify
% missing phrases, additional members, and duplicate assignments
% for correction.

Each final category contains its name, definition,
inclusion and exclusion boundaries, and its original member
phrases. Then the resulting taxonomies relabel all extracted
chains. Metrics are computed from these
normalized category sequences.

\subsection{Metric Computation}

Metrics are computed programmatically from the relabeled
chains. Here, an $n$-gram denotes slots within one causal
chain.
Given $N>0$ chains, let $U_4$, $U_{CMO}$, and $U_{MOR}$
denote the numbers of distinct four-slot patterns,
C--M--O triples, and M--O--R triples, respectively. Then
\begin{equation}
\begin{aligned}
\mathrm{Rep}_4 &= 1-\frac{U_4}{N},\\
\mathrm{Rep}_3 &= \frac{1}{2}\left[
\left(1-\frac{U_{CMO}}{N}\right)+
\left(1-\frac{U_{MOR}}{N}\right)\right].
\end{aligned}
\end{equation}

For the normalized conditional entropy defined in the
main text, the reference support threshold is $\tau=3$.
The reported entropy is the arithmetic mean
of the normalized summaries for $M\mid C$, $O\mid M$,
and $R\mid O$.

%% file: appendices/07.tex
\section{Additional Experimental Details and Results}
\label{app:additional-experiments}

\begin{table*}[!htbp]
  \centering
  \small
  \setlength{\tabcolsep}{1.5pt}
  \renewcommand{\arraystretch}{1.15}

\resizebox{\textwidth}{!}{%
\begin{tabular}{@{}cl*{17}{c}@{}}
    \toprule
    & & \multicolumn{8}{c}{\textbf{Script Logic}}
    & \multicolumn{6}{c}{\textbf{Script Quality}}
    & \multicolumn{3}{c}{\textbf{Homogeneity}} \\
    \cmidrule(lr){3-10}
    \cmidrule(lr){11-16}
    \cmidrule(lr){17-19}

    \textbf{Model} & \textbf{Method}
    & \textbf{CC} & \textbf{CS} & \textbf{KS}
    & \textbf{RS} & \textbf{RST} & \textbf{SP}
    & \textbf{RD} & \textbf{Avg}
    & \textbf{CQ} & \textbf{DE} & \textbf{PP}
    & \textbf{ES} & \textbf{SQ} & \textbf{Avg}
    & \textbf{4G}$\downarrow$
    & \textbf{3G}$\downarrow$
    & $\overline{H}\uparrow$ \\
    \midrule

    % ===== DeepSeek-V4-Pro =====
    \multirow{3}{*}{\shortstack[c]{DeepSeek-\\V4-Pro}}
    & \methodcell{DOME}
    & \scorecell{75.5}{1.1}
    & \scorecell{69.0}{1.0}
    & \scorecell{78.3}{0.2}
    & \scorecell{76.3}{0.5}
    & \scorecell{72.5}{0.7}
    & \scorecell{78.5}{0.5}
    & \scorecell{75.6}{1.3}
    & \scorecell{75.1}{0.4}
    & \scorecell{87.5}{0.7}
    & \scorecell{86.2}{0.5}
    & \scorecell{79.0}{1.0}
    & \scorecell{83.4}{1.1}
    & \scorecell{87.2}{0.1}
    & \scorecell{84.7}{0.5}
    & \scorecell{8.23}{0.70}
    & \scorecell{17.2}{0.43}
    & \scorecell{0.853}{0.012} \\[1.3pt]

    & \methodcell{StoryWriter}
    & \scorecell{74.2}{0.8}
    & \scorecell{69.4}{1.1}
    & \scorecell{78.2}{0.3}
    & \scorecell{75.0}{0.8}
    & \scorecell{72.1}{0.5}
    & \scorecell{76.5}{0.8}
    & \scorecell{67.8}{1.1}
    & \scorecell{73.3}{0.7}
    & \scorecell{87.7}{1.3}
    & \scorecell{86.0}{0.8}
    & \scorecell{77.2}{0.6}
    & \scorecell{85.9}{0.3}
    & \scorecell{86.7}{0.3}
    & \scorecell{84.7}{0.6}
    & \scorecell{6.74}{0.64}
    & \scorecell{14.5}{1.0}
    & \scorecell{0.829}{0.0068} \\[1.3pt]

    & \methodcell{Ours}
    & \scorecell{86.0}{0.1}
    & \scorecell{85.7}{0.5}
    & \scorecell{85.0}{0.4}
    & \scorecell{83.6}{0.5}
    & \scorecell{85.4}{0.4}
    & \scorecell{86.9}{0.2}
    & \scorecell{89.8}{0.1}
    & \bestcell{86.1}{0.2}
    & \scorecell{92.7}{0.1}
    & \scorecell{91.1}{0.6}
    & \scorecell{87.8}{0.1}
    & \scorecell{89.9}{0.3}
    & \scorecell{89.7}{0.1}
    & \bestcell{90.2}{0.2}
    & \bestcell{3.21}{0.57}
    & \bestcell{10.2}{0.51}
    & \bestcell{0.915}{0.0030} \\

    \midrule

    % ===== GLM5.3-Flash =====
    \multirow{3}{*}{\shortstack[c]{GLM5.3-\\Flash}}
    & \methodcell{DOME}
    & \scorecell{83.9}{0.8}
    & \scorecell{80.2}{0.6}
    & \scorecell{83.8}{0.3}
    & \scorecell{82.2}{0.7}
    & \scorecell{81.7}{0.6}
    & \scorecell{84.9}{0.7}
    & \scorecell{86.5}{0.7}
    & \scorecell{83.3}{0.4}
    & \scorecell{92.0}{0.8}
    & \scorecell{90.1}{0.3}
    & \scorecell{86.4}{0.7}
    & \scorecell{87.0}{0.6}
    & \scorecell{89.4}{0.1}
    & \scorecell{89.0}{0.3}
    & \scorecell{8.03}{0.94}
    & \scorecell{15.1}{0.94}
    & \scorecell{0.881}{0.0062} \\[1.3pt]

    & \methodcell{StoryWriter}
    & \scorecell{85.1}{0.5}
    & \scorecell{83.8}{0.5}
    & \scorecell{84.7}{0.6}
    & \scorecell{84.2}{0.3}
    & \scorecell{83.2}{0.6}
    & \scorecell{86.7}{0.2}
    & \scorecell{85.9}{0.7}
    & \scorecell{84.8}{0.4}
    & \scorecell{92.4}{0.3}
    & \scorecell{89.9}{0.1}
    & \scorecell{85.0}{0.4}
    & \scorecell{86.5}{0.7}
    & \scorecell{89.1}{0.3}
    & \scorecell{88.6}{0.2}
    & \scorecell{7.08}{1.1}
    & \scorecell{14.9}{0.90}
    & \scorecell{0.870}{0.010} \\[1.3pt]

    & \methodcell{Ours}
    & \scorecell{89.7}{0.3}
    & \scorecell{89.5}{0.3}
    & \scorecell{89.1}{0.4}
    & \scorecell{88.5}{0.1}
    & \scorecell{88.8}{0.7}
    & \scorecell{90.5}{0.3}
    & \scorecell{90.9}{0.1}
    & \bestcell{89.6}{0.2}
    & \scorecell{95.1}{0.1}
    & \scorecell{93.0}{0.1}
    & \scorecell{91.6}{0.4}
    & \scorecell{91.8}{0.6}
    & \scorecell{90.9}{0.1}
    & \bestcell{92.5}{0.0}
    & \bestcell{2.30}{0.19}
    & \bestcell{9.13}{0.77}
    & \bestcell{0.927}{0.0050} \\

    \bottomrule
\end{tabular}
}

\caption{
Cross-model generalization results on 40-episode
script generation using DeepSeek-V4-Pro and
GLM5.3-Flash.
Results are reported as mean $\pm$ standard deviation.
Best average and homogeneity scores within each
model group are highlighted in bold.
}
\label{tab:cross-model-results}
% \end{table*}

% \begin{table*}[!htbp]
%   \centering
%   \small
%   \setlength{\tabcolsep}{1.5pt}
%   \renewcommand{\arraystretch}{1.15}
\par\vspace{10pt}

\resizebox{\textwidth}{!}{%
\begin{tabular}{@{}l*{17}{c}@{}}
    \toprule
    & \multicolumn{8}{c}{\textbf{Script Logic}}
    & \multicolumn{6}{c}{\textbf{Script Quality}}
    & \multicolumn{3}{c}{\textbf{Homogeneity}} \\
    \cmidrule(lr){2-9}
    \cmidrule(lr){10-15}
    \cmidrule(lr){16-18}

    \textbf{Method}
    & \textbf{CC} & \textbf{CS} & \textbf{KS}
    & \textbf{RS} & \textbf{RST} & \textbf{SP}
    & \textbf{RD} & \textbf{Avg}
    & \textbf{CQ} & \textbf{DE} & \textbf{PP}
    & \textbf{ES} & \textbf{SQ} & \textbf{Avg}
    & \textbf{4G}$\downarrow$
    & \textbf{3G}$\downarrow$
    & $\overline{H}\uparrow$ \\
    \midrule

    % ===== Ours =====
    \methodcell{Ours}
    & \scorecell{80.4}{0.6}
    & \scorecell{79.2}{0.3}
    & \scorecell{81.8}{0.5}
    & \scorecell{78.0}{1.2}
    & \scorecell{79.0}{0.6}
    & \scorecell{83.7}{0.7}
    & \scorecell{87.3}{0.9}
    & \bestcell{81.3}{0.7}
    & \scorecell{88.5}{0.3}
    & \scorecell{88.1}{0.6}
    & \scorecell{82.4}{0.8}
    & \scorecell{85.0}{0.4}
    & \scorecell{87.7}{0.1}
    & \bestcell{86.3}{0.4}
    & \bestcell{3.08}{0.41}
    & \bestcell{9.22}{0.63}
    & \bestcell{0.929}{0.0056} \\[0.5pt]

    % ===== w/o Future-Aware Memory =====
    \methodcell{w/o Future-Aware Mem}
    & \scorecell{74.0}{0.3}
    & \scorecell{74.5}{1.3}
    & \scorecell{79.5}{0.6}
    & \scorecell{73.0}{0.3}
    & \scorecell{73.2}{0.4}
    & \scorecell{80.0}{0.1}
    & \scorecell{67.3}{1.4}
    & \scorecell{74.5}{0.3}
    & \scorecell{79.6}{0.8}
    & \scorecell{85.0}{0.1}
    & \scorecell{78.0}{0.4}
    & \scorecell{77.2}{0.8}
    & \scorecell{86.4}{0.5}
    & \scorecell{81.2}{0.3}
    & \scorecell{4.65}{0.41}
    & \scorecell{12.2}{0.86}
    & \scorecell{0.914}{0.0084} \\[0.5pt]

    % ===== w/o Memory & Tree Search =====
    \methodcell{w/o Future-Aware Mem\\
\& Tree Search}
    & \scorecell{72.1}{0.8}
    & \scorecell{69.5}{0.8}
    & \scorecell{77.4}{0.8}
    & \scorecell{71.8}{0.5}
    & \scorecell{70.4}{0.5}
    & \scorecell{77.3}{0.9}
    & \scorecell{66.2}{0.4}
    & \scorecell{72.1}{0.3}
    & \scorecell{83.7}{0.4}
    & \scorecell{82.7}{0.2}
    & \scorecell{72.6}{0.5}
    & \scorecell{80.3}{0.6}
    & \scorecell{86.7}{0.4}
    & \scorecell{81.2}{0.4}
    & \scorecell{7.13}{1.2}
    & \scorecell{16.4}{0.81}
    & \scorecell{0.887}{0.0091} \\[0.5pt]

    % ===== w/o Agent Pipeline (Direct) =====
    \methodcell{w/o Multi-Agent Pipeline\\
(Direct Generation)}
    & \scorecell{68.7}{0.7}
    & \scorecell{63.5}{0.7}
    & \scorecell{74.4}{0.2}
    & \scorecell{66.5}{0.5}
    & \scorecell{66.9}{0.7}
    & \scorecell{70.4}{0.2}
    & \scorecell{81.8}{0.4}
    & \scorecell{70.3}{0.1}
    & \scorecell{75.9}{0.6}
    & \scorecell{80.1}{0.3}
    & \scorecell{71.4}{0.2}
    & \scorecell{75.5}{0.9}
    & \scorecell{81.3}{0.8}
    & \scorecell{76.8}{0.2}
    & \scorecell{6.87}{0.14}
    & \scorecell{16.8}{0.68}
    & \scorecell{0.887}{0.012} \\

    \bottomrule
\end{tabular}
}

\caption{
Ablation results on 40-episode script generation
using Qwen3.6-27B.
Starting from Ours, we progressively remove
Future-Aware Memory and Evolutionary Tree Search
for Outline Refinement, while retaining the original
multi-Agent Pipeline.
The Direct baseline generates scripts without
the multi-Agent Pipeline.
}
\label{tab:ablation-results-full}
\end{table*}

Tables~\ref{tab:cross-model-results}
and~\ref{tab:ablation-results-full} provide the complete
cross-model and ablation results corresponding to
Tables~\ref{tab:cross-model-results-average}
and~\ref{tab:ablation-results}, respectively.
For space considerations, the main-text tables report only
the aggregate Logic and Quality scores alongside the
homogeneity metrics. The full tables additionally report
all seven Logic and five Quality categories, providing
a finer-grained view of performance across evaluation criteria.
Both experiments use 40-episode scripts under the settings
described in the main text. Results are reported as means
and sample standard deviations across three runs.